\documentclass[10pt,twocolumn,letterpaper]{article}

\usepackage{iccv}
\usepackage{times}
\usepackage{epsfig}
\usepackage{graphicx}
\usepackage{amsmath}
\usepackage{amssymb}
\usepackage{multirow}

\usepackage[pagebackref=true,breaklinks=true,letterpaper=true,colorlinks,bookmarks=false]{hyperref}
\hypersetup{
	plainpages=false,
	colorlinks=true,              %
	linkcolor=blue,               %
	anchorcolor=blue,             %
	citecolor=blue,               %
	filecolor=blue,               %
	urlcolor=blue,                %
	pdfview=FitH,                 %
	pdfstartview=FitH,            %
	pdfpagelayout=SinglePage      %
}

\newcommand{\sg}{ \texttt{sg}}

\newcommand{\tablestyle}[2]{\setlength{\tabcolsep}{#1}\renewcommand{\arraystretch}{#2}\centering\footnotesize}
\newcommand{\omitme}[1]{}

\usepackage{url}            %
\usepackage{booktabs}       %
\usepackage{amsfonts}       %
\usepackage{nicefrac}       %
\usepackage{microtype}      %
\usepackage{graphicx}
\usepackage{xcolor}
\usepackage[numbers,sort]{natbib} %
\usepackage{wrapfig}
\usepackage{xspace}
\usepackage{stmaryrd}

\newcommand\added[1]{\textcolor{black}{#1}}

\definecolor{grey}{rgb}{0.5, 0.5, 0.5}
\newcommand\spv[1]{\textcolor{grey}{#1}}

\newcommand{\algo}{\texttt{\textbf{BraVe}}\xspace}

\newcommand{\targetview}{broad view\xspace}
\newcommand{\sourceview}{narrow view\xspace}
\newcommand{\targetviews}{broad views\xspace}

\newcommand{\bton}{b\rightarrow n}

\usepackage{pifont}
\newcommand{\cmark}{\ding{51}}%
\newcommand{\xmark}{\ding{55}}%

\newcommand{\xF}{x_n}
\newcommand{\xB}{x_b}

\newcommand{\fF}{f_{n}}
\newcommand{\fB}{f_{b}}
\newcommand{\pF}{g}
\newcommand{\pB}{g}
\newcommand{\zF}{z_n}
\newcommand{\zB}{z_b}
\newcommand{\qF}{h}
\newcommand{\qB}{h}

\let\originalleft\left
\let\originalright\right
\renewcommand{\left}{\mathopen{}\mathclose\bgroup\originalleft}
\renewcommand{\right}{\aftergroup\egroup\originalright}

\usepackage[pagebackref=true,breaklinks=true,letterpaper=true,colorlinks,bookmarks=false]{hyperref}

\iccvfinalcopy 

\def\iccvPaperID{****} 

\begin{document}

\title{Broaden Your Views for Self-Supervised Video Learning \vspace{-10pt}}
\author{
	\renewcommand{\thefootnote}{\fnsymbol{footnote}}
	Adri\`a Recasens\textsuperscript{1}\footnotemark[2]
	\renewcommand*{\thefootnote}{\arabic{footnote}}
	\quad
	Pauline Luc\textsuperscript{1}
	\quad
	Jean-Baptiste Alayrac\textsuperscript{1}
	\quad
	Luyu Wang\textsuperscript{1}
	\quad
	Ross Hemsley\textsuperscript{1}
	\\
	Florian Strub\textsuperscript{1}
	\quad
	Corentin Tallec\textsuperscript{1}
	\quad
	Mateusz Malinowski\textsuperscript{1}
	\quad
	Viorica P{\u a}tr{\u a}ucean\textsuperscript{1}
	\quad
	Florent Altché\textsuperscript{1}
	\\
	Michal Valko\textsuperscript{1}
	\quad
	Jean-Bastien Grill\textsuperscript{1}
	\quad
	A\"aron van den Oord\textsuperscript{1}
	\quad
	Andrew Zisserman\textsuperscript{1,2}
	\\
	$^1$\small DeepMind \quad  $^2$VGG, Dept.\  of Engineering Science, University of Oxford \vspace{6pt}
	\\
       \normalsize \url{http://github.com/deepmind/brave}
}

\maketitle
\ificcvfinal\thispagestyle{empty}\fi

\begin{abstract}%
Most successful self-supervised learning methods are trained to align the representations of two independent views from the data.
State-of-the-art methods in video are inspired by image techniques, where these two views are similarly extracted by cropping and augmenting the resulting crop.
However, these methods miss a crucial element in the video domain: time.
We introduce BraVe, a self-supervised learning framework for video. 
In BraVe, one of the views has access to a narrow temporal window of the video while the other view has a broad access to the video content. 
Our models learn to generalise from the narrow view to the general content of the video.
Furthermore, BraVe processes the views with different backbones, enabling the use of alternative augmentations or modalities into the broad view such as optical flow, randomly convolved RGB frames, audio or their combinations.
We demonstrate that BraVe achieves state-of-the-art results in self-supervised representation learning
on standard video and audio classification benchmarks including UCF101, HMDB51, Kinetics,
ESC-50 and AudioSet.
\vspace*{-0.5cm}
\end{abstract}

\section{Introduction}
	\renewcommand{\thefootnote}{\fnsymbol{footnote}}
	\footnotetext[2]{Correspondence to: Adrià Recasens (arecasens@google.com)}
	\renewcommand*{\thefootnote}{\arabic{footnote}}

Over the past few years, self-supervised methods have revolutionized the field of representation learning~\cite{chen2020simple,henaff2019data,richemond2020byol}.
These methods directly learn from data without the need for manually defined labels that are hard to get at scale.
Doing so, one can successfully leverage large amounts of \emph{uncurated} data to improve representations.
Even more importantly, self-supervised learning enables richer training tasks to be defined, compared to the standard approach of trying to categorize diverse visual inputs into a fixed set of categories.
This has led to self-supervised representations outperforming supervised ones on downstream tasks~\cite{he2020momentum}.
Video is a natural 
domain for self-supervised learning since data is rich and abundant but hard to annotate at scale due to the additional temporal complexity.
However, most methods in the video domain take direct inspiration from methods developed for images without fully taking advantage of its distinctly different dimension: time.

\begin{figure}[t!]
	\centering
	\includegraphics[width=0.48\textwidth]{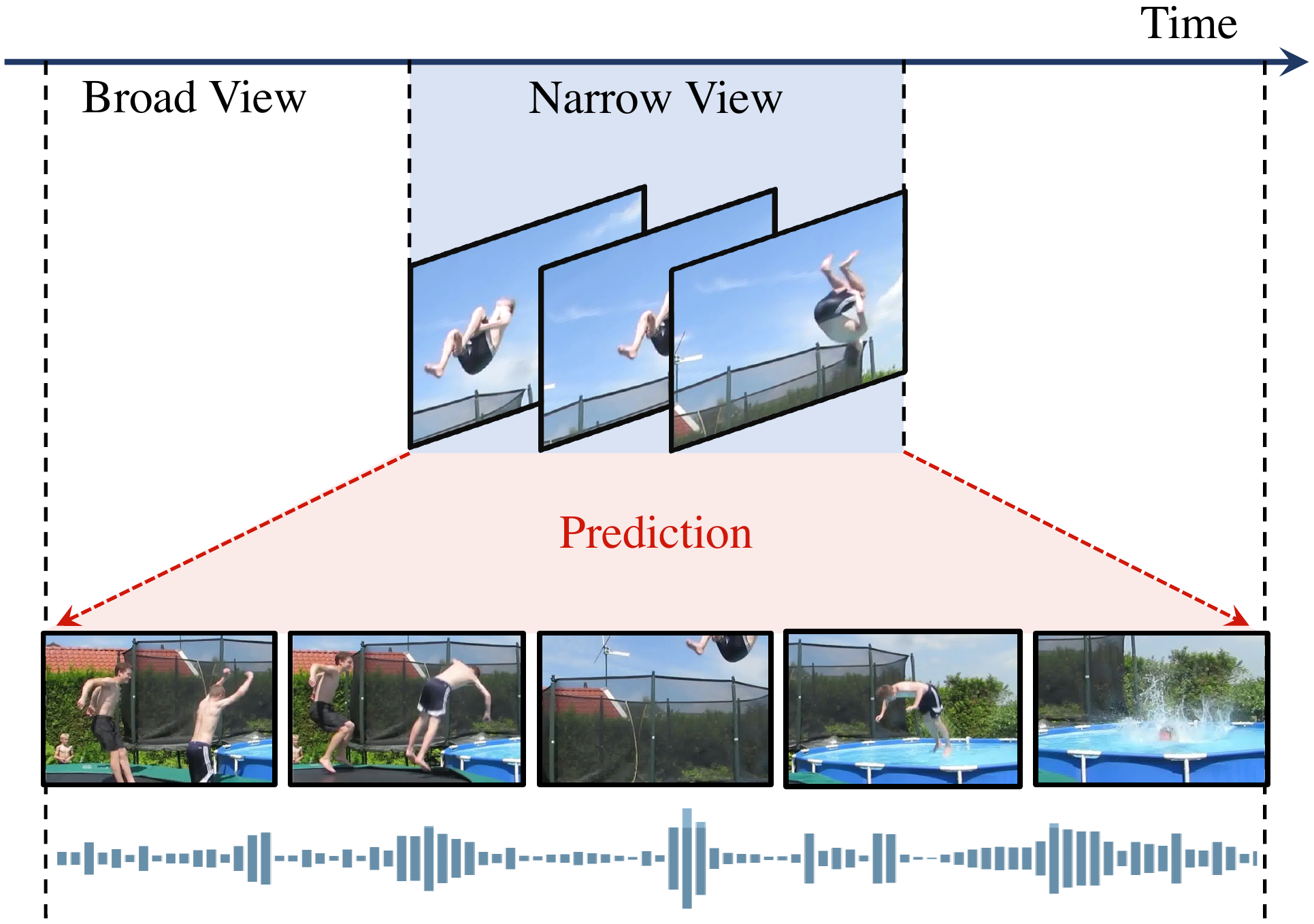}
	\vspace*{-0.05cm}
	\caption{\small 
		Given a \emph{\sourceview} corresponding to a video clip of a few seconds, \algo is tasked with predicting a \emph{\targetview} that spans a longer temporal context of the video in different modalities (here visual and audio).
		Solving that task requires the representation to extrapolate what happened before, during and after the \emph{\sourceview}, and results in state-of-the-art video representations.
	} 
	\vspace*{-0.3cm}
	\label{fig:teaser}
\end{figure}

In particular, one common aspect of self-supervised methods for images is to extract two views from a given instance using the same general augmentation procedure, feed them into a shared backbone, and extract a supervisory signal from the fact that these two views originate from the same source.
This is true for most recent approaches irrespective of their underlying learning principle: contrastive approaches~\cite{chen2020simple}, clustering-based method~\cite{caron2018DeepCF}, or regression algorithms~\cite{richemond2020byol}.
The same principle has been followed in the video domain~\cite{alayrac2020self,qian2020spatiotemporal}.
Specifically, most video methods extract the different views from a source video clip in a \emph{symmetric} fashion with respect to time: all extracted views have the same temporal extent in the video~\cite{alayrac2020self,korbar2018cooperative,qian2020spatiotemporal, feichtenhofer2021large}.
However, doing so does not benefit from learning from information contained at different time scales.

In this paper, we introduce an algorithm dubbed ``\textbf{Br}o\textbf{a}den your \textbf{V}i\textbf{e}ws'' (\algo), that breaks this symmetry in order to improve representation learning from videos.
In detail, given a \emph{\sourceview} corresponding to a video clip of a few seconds, \algo learns a representation by predicting a \emph{\targetview}  that spans the longer temporal context of the full video clip as illustrated in Figure~\ref{fig:teaser}.
Solving such a task requires extrapolating
to the general context in which a given event occurs.
In the example of Figure~\ref{fig:teaser}, one has to predict what happened before the person is in the sky (they probably jumped with the help of some device, given the height), as well as what is going to happen next (they will probably fall down somewhere soft) in order to solve the task.
This task arguably requires a good understanding of the structure of events and is therefore a promising task for learning representations.
While related local-to-global proxy tasks have been studied in the image domain via network architectural designs~\cite{hjelm2018learning,AMDIM} or multi-size cropping~\cite{chen2020simple}, applying these techniques to videos is not straightforward, because of the increased computational complexity incurred by the time dimension and the artifacts introduced when doing similar resize operations in spatio-temporal volumes.
To address this challenge, we propose to process \targetviews with a dedicated model.
We demonstrate that under a fixed computational budget, learning from the supervision provided by our \targetviews performs better than alternatives relying on symmetric augmentation procedures.
Our algorithm is simple and does not require a cumbersome creation of explicit negatives as in contrastive methods.
Instead we use a direct regression-based approach inspired by BYOL~\cite{grill2020bootstrap}, where the views are processed by dedicated backbones and regress each other.
Breaking the symmetry enables the use of stronger augmentations and different modalities for the \targetview, which improves the quality of the final representations.

\paragraph{Contributions.}
We make the following contributions: ~~~
\textbf{(i)} We propose a novel framework for representation learning, called \algo, which generates views at different time scales and learns representations via simple regression across views,
\textbf{(ii)} We explore using different augmentations and modalities in the broad view such as audio, flow or randomly convolved RGB frames. 
\textbf{(iii)} We evaluate this framework in the video domain, both with and without audio as an auxiliary supervisory signal, where we obtain \emph{state-of-the-art} results on video and audio classification benchmarks UCF101, HMDB51, Kinetics, ESC-50 and AudioSet.

\section{Related work}
\label{sec:related}

\noindent
\textbf{Image-based self-supervised learning.}
Most successful self-supervised methods
learn a representation by defining a \textit{pretext task}, whose resolution typically entails learning useful representations
~\cite{gidaris2018unsupervised,zhang2016colorful,noroozi2016unsupervised,pathak2016context,caron2018DeepCF,caron2020unsupervised,doersch2015unsupervised,doersch2017multi}.
In particular, contrastive methods have provided spectacular performance~\cite{dosovitskiy2014discriminative,chen2020simple,he2020momentum,MOCOv2,tian2019contrastive,henaff2019data,bachman2019learning,tian2020makes,PIRL,le2020contrastive}.
Contrastive methods learn by pulling representations of different transformations of the same image (positive instances) closer, and pushing representations of different images (negatives) apart~\cite{oord2018representation,bachman2019learning}. 
The main drawbacks of contrastive approaches are that they require a careful choice of positive and negative pairs~\cite{tian2020makes} and that they often rely on large number of such negatives, inducing a high computational cost~\cite{chen2020simple}.
Alternatives to the contrastive approach, such as clustering and regression, avoid the need and cost of multiple negatives.
Clustering-based methods~\cite{tian2017deepcluster,alwassel2019self,asano2020selflabelling,bautista2016cliquecnn,caron2018DeepCF,caron2020unsupervised,huang2019unsupervised,xie2016unsupervised} alternate between  learning representations using clusters as targets, and clustering using the current representations (either online or offline).
Most related to our work are regression-based methods that instead try to directly regress a representation extracted from a different view of the image~\cite{gidaris2020learning,richemond2020byol}.
\algo is directly inspired from~\cite{grill2020bootstrap} but the views come from different modalities and augmentations, are processed by dedicated backbones and regress each other.

	\begin{figure*}[t!]
		\centering
		\includegraphics[width=\textwidth]{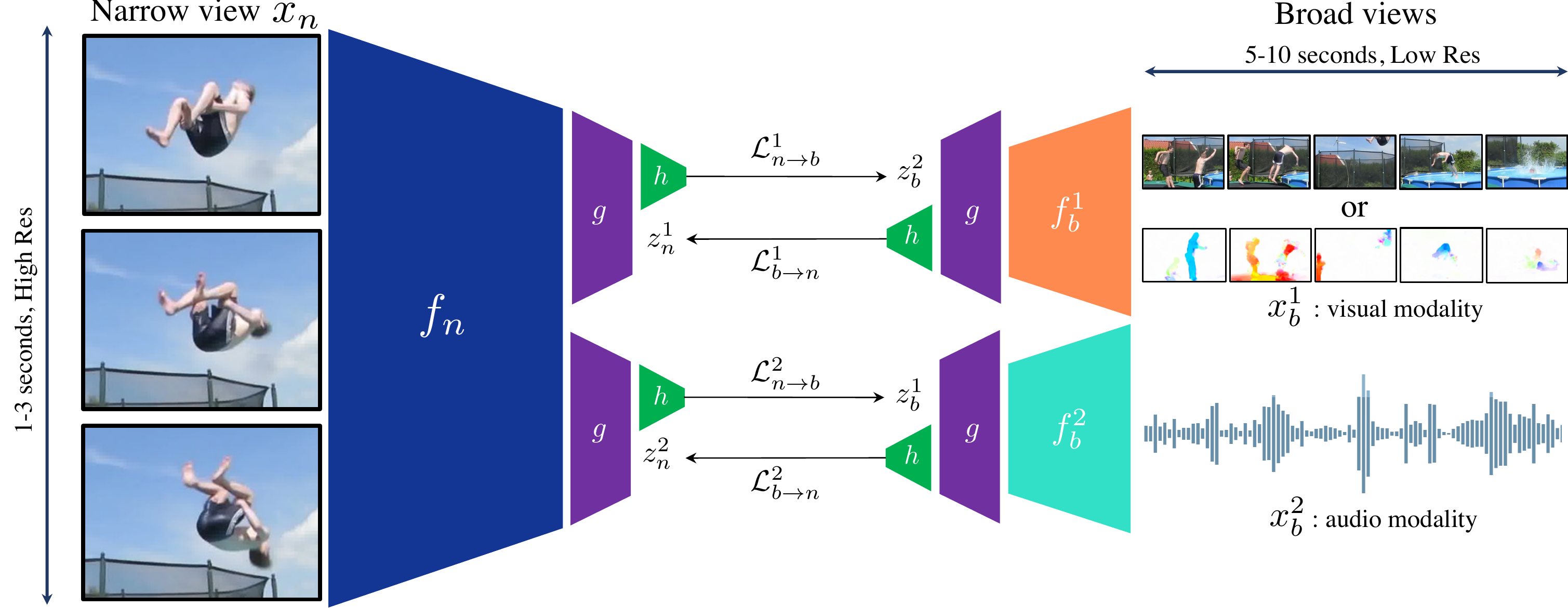}
		\caption{\algo. Given a \sourceview $x_n$ spanning a few seconds at high resolution and broad views $x_b^1$ and $x_b^2$ covering a larger temporal extent in the video for different modalities, we train independent networks running on the narrow and the broad views to mutually regress each other. This is done by defining two regression losses: $\mathcal{L}_{n\rightarrow b}$ to predict a broad view from the narrow view, and $\mathcal{L}_{\bton}$ to enforce the other way around. To avoid collapse of the learned representations, we introduce three stages of processing as previously done in BYOL~\cite{grill2020bootstrap}: backbone networks ($\fF$ for the narrow view and $f_b^1,f_b^2$ for the broad views), projector network $\pF$ and predictor network $\qF$.
		For the broad views, we consider both visual (RGB frames or optical flow) and audio modalities (spectograms).}
		\label{fig:model_figure}
		\vspace*{-0.05cm}
	\end{figure*}

\noindent
\textbf{Video-based self-supervised learning.}
In the video domain, the pretext tasks for self-supervision have included predicting the future in pixel space by minimising an MSE loss~\cite{vondrick2018tracking,pmlr-v37-srivastava15,PatrauceanHC16} or adversarial losses~\cite{Vondrick16a,mathieu2015deep}. 
However, the predictions of these models are usually blurred and cannot go beyond predicting short clips into the future.
To avoid these difficulties, other works focus on learning representations in a more abstract space, by using pretext tasks that predict the temporal order of video frames~\cite{misra2016shuffle} or the arrow of time~\cite{Wei_2018_CVPR}.
In this direction also, video contrastive methods have been very successful~\cite{qian2020spatiotemporal,Han20,MEMDPC,hjelm2020representation}. 
In addition to data augmentations used for images, these works use temporal cues to build \textit{positive pairs}. Yet the costs of training such systems are significant and complex hard-negative mining strategies are needed to improve the training efficiency~\cite{faghri2017vse}. 
\added{Concurrent to our work, \cite{feichtenhofer2021large} introduces $\rho$BYOL which consists in directly applying BYOL, an image self-supervised technique, to video. Although $\rho$BYOL circumvents the use of negatives, it still requires an EMA (Exponential Moving Average) network to generate the targets, increasing the computational complexity. Our method avoids this computational overhead while obtaining state-of-the-art performance on popular video benchmarks.}
Furthermore, our approach may leverage predictive tasks, such as predicting other crops in the video or optical flow,
reminiscent of earlier predictive work~\cite{pmlr-v37-srivastava15,walker2016uncertain};
but predicting in a learned feature space by building on a more recent self-supervised approach~\cite{grill2020bootstrap}.

\noindent
\textbf{Audio-video self-supervised learning.}
Video and audio have been used as a rich source of self-supervision~\cite{arandjelovic17look,arandjelovic2018objects,Senocak_2018_CVPR,owens2018audio,korbar2018cooperative,alwassel2019self,mandela2020datatrans,morgado20avid}. 
A simple but effective approach to train representations consists in classifying whether a video clip and an audio sample correspond to each other \cite{arandjelovic17look,arandjelovic2018objects,Senocak_2018_CVPR,owens2018audio,korbar2018cooperative}.
Some works propose to use language obtained from speech recognition as an additional supervisory signal~\cite{alayrac16unsupervised,alayrac2020self,malmaud15what,miech19howto100m,miech2019end2end,Sener_2015_ICCV,sigurdsson2020visual,sun2019videobert}. 
Related to ours, recent work finds that distilling flow and audio into a RGB encoder leads to strong representations~\cite{piergiovanni2020evolving}, using an evolutionary search algorithm on the loss function.
In contrast with this approach, our framework does not require defining modality-specific losses, is simpler to train (no need to balance the losses), and obtains better performance across the board.


\section{Broaden Your Views for Self-Supervised Video Learning}
	\label{sec:approach}\noindent
	In this section, we detail our approach dubbed \algo for learning self-supervised state-of-the-art representations from a large set of videos, as measured by performance when transferring to downstream tasks.
	\algo, illustrated in Figure~\ref{fig:model_figure}, learns by direct regression from a high resolution narrow view that only spans a short clip to a lower resolution broader view which covers a larger temporal context of the video. 
	Multiple options can be considered for the broad view: it can either come from the same modality as the narrow view (RGB in our case) or a different one such as  flow or audio.
	Multiple views can also be combined to further improve performance.
	Next, we formally describe the learning framework in Section~\ref{subsect:framework} and provide intuition why this may be a good self-supervised objective.
	Then, in Section~\ref{subsect:multimodal}, we describe the components and views we use in practice in two standard settings: learning from (i) visual signals alone, and from (ii) visual and audio modalities.
	
	\subsection{The \algo learning framework}
	\label{subsect:framework}\noindent
	\textbf{General overview.}  
	Given a video $x$ that can be composed of multiple modalities, we randomly extract two complementary views: a \sourceview $x_n$ that spans a short timeframe in the video (around 1-3 seconds) and a \targetview $x_b$ that covers a larger extent of the video (around 5-10 seconds).
	Details on how these views are obtained are given in Section~\ref{subsect:multimodal}.
	By introducing this temporal asymmetry in the creation of the views, the proposed task consists in extrapolating the full context of the video (the \targetview) from only a small portion of the video (the \sourceview) as illustrated in Figure~\ref{fig:teaser}.
	We hypothesize that to solve this task, good representations must be learned, which can then be useful for semantic downstream tasks.
	More formally, we train networks to minimize the training loss $\mathcal{L}$ defined for a given video $x$ as follows:
	\begin{align}
		\label{eq:main}
		 \mathcal{L}(x)=\underbrace{\mathcal{L}_{n\rightarrow b}(x)}_{\text{Narrow}\rightarrow\text{Broad}} + \underbrace{\mathcal{L}_{\bton}(x)}_{\text{Broad}\rightarrow \text{Narrow}}.
	\end{align}
	This loss is composed of two terms: (i) a prediction loss from the narrow to the broad view, and (ii) a complementary loss to regress the narrow view from the broad view.

	\noindent
	\textbf{\algo: losses and architectures.}
	For simplicity and computational purposes, we opt for simple regression losses for $\mathcal{L}_{n\rightarrow b}$ and $\mathcal{L}_{\bton}$.
	This is indeed simpler than standard contrastive losses that require large batches and therefore high compute to work well~\cite{chen2020simple}.
	One challenge however, is the risk of collapse, since a trivial solution could be to always predict a constant which would lead to perfect regression losses across views.
	To avoid this, we draw inspiration from recent work~\cite{guo2020pbl,grill2020bootstrap} in the way we design our networks and losses, as detailed next.
	
	\noindent As illustrated in Figure~\ref{fig:model_figure}, we first define a backbone network $\fF$ whose role is to extract a representation from the \sourceview $\xF$.
	Similarly, we define a backbone network $\fB$ acting on the \targetview $\xB$.
	Note that in our framework, the parameters and even the underlying architectures of $\fF$ and $\fB$ can differ since they act on views of a different nature.
	Secondly, we define a single projector $\pF$, which is used to transform the representations of the different views projecting $\fF(\xF)$ and $\fB(\xB)$ to yield the narrow embedding $\zF=\pF(\fF(\xF))$ and the broad one $\zB=\pB(\fB(\xB))$. Note that the projectors are shared across all the views, which we empirically observe to improve performance. Whenever the output dimensions of $\fF$ and $\fB$ do not match, we apply a single linear layer on $\fB(\xB)$ to match the dimensionality of both tensors. 
	Similarly, inspired by~\cite{grill2020bootstrap}, we then define a third stage of processing using a predictor $\qF$ that takes the projected embeddings $\zF$ and $\zB$ and computes predictions $\qF(\zF)$ and $\qB(\zB)$. Prediction $\qF(\zF)$ is used to regress the \targetview $\zB$ using the following loss:
	\begin{align}
	\label{eq:loss_ntob}
		\mathcal{L}_{n\rightarrow b}(x) = \left\| \frac{\qF(\zF)}{\| \qF(\zF) \|_2} -\sg\left[\frac{\zB}{\| \zB \|_2}\right]  \right\|_2^2,
	\end{align}
	where $\sg[ \cdot ]$ denotes the ``stop gradient'' operator, which operates on its input as the identity, but has zero partial derivatives.
	Since the loss $\mathcal{L}_{n\rightarrow b}$ only depends on the networks associated with the \sourceview, we also define a loss to provide training signal for the \targetview network.
	To that end, we use prediction $\qB(\zB)$ to regress the \sourceview embedding $\zF$ using the following loss:
	\begin{align}
		\mathcal{L}_{\bton}(x) = \left\| \frac{\qB(\zB)}{\| \qB(\zB) \|_2} - \sg\left[\frac{\zF}{\| \zF \|_2}\right]\right\|_2^2.
	\end{align}
	
	The role of the predictor is crucial to avoid collapse as found in~\cite{grill2020bootstrap}, which we confirm experimentally. The same is true for the stop gradient operator.
	Differently from~\cite{grill2020bootstrap}, we do not use exponential moving averages (EMA) on the weights of the network that process the view  being regressed.
	Unlike~\cite{grill2020bootstrap, feichtenhofer2021large}, who required the moving average for improved performance, we find that this is not necessary in our case. 

	\noindent 
	\textbf{Intuitions about what needs to be learned by \algo.}
	While the proposed approach avoids plain collapse of the representations, it is also important to question what needs to be learned in order for the loss~\eqref{eq:main} to be optimized.
	In particular, we want the narrow backbone to learn to predict the full context represented by the \targetview.
	However, one challenge is to prevent the broad backbone from instead simply learning to throw the broad information away and only keeping the signal contained in the \sourceview.
	To avoid this, we sample the narrow and broad views independently in time when they come from the same visual modality so that it is difficult for the broad backbone to predict what the \sourceview is going to be.
	By doing so, we argue that the best solution to solve the task is for the narrow backbone to extrapolate what is happening in the broad view.
	We empirically verify the importance of this independent sampling in our experiments in section~\ref{sec:results}.
	
	\noindent
	\added{
	\textbf{Dealing with multiple views from one modality.}
	\algo can be extended to handle $K$ broad views (with $K>1$) coming from the same modality.
	To do so, we keep a single backbone $\fB$ for all broad views. 
	For each broad view $x^k_{b}$, we individually compute the projection $z_b^k = g(\fB(x^k_{b}))$ and the prediction $h(z_b^k)$. 
	The target for the narrow backbone used in $\mathcal{L}_{n\rightarrow b}$~\eqref{eq:loss_ntob} is obtained by averaging these broad views projections:
	$z_b = \frac{1}{K}\sum z_b^k$.
	To compute the $\mathcal{L}_{b\rightarrow n}$ loss, we average the individual losses $\mathcal{L}_{\bton}^k$ of the different predictions $h(z_b^k)$:
	\begin{align}
		\mathcal{L}_{\bton}^k(x) = \left\| \frac{\qB(z_b^k)}{\| \qB(z_b^k) \|_2} - \sg\left[\frac{\zF}{\| \zF \|_2}\right]\right\|_2^2.
	\end{align}}
	This setting has many similarities with $\rho$BYOL~\cite{feichtenhofer2021large}. However, instead of using Exponential Moving Average, we use a different backbone for the broad view, which reduces the computation (dividing by half the number of forward passes) and enables the use of other modalities such as flow or audio.

	\noindent 
	\textbf{Dealing with multiple views from different modalities.}
	\algo can also be extended to handle $K$ broad views (with $K>1$) coming from different modalities.
	To do so and as illustrated in Figure~\ref{fig:model_figure}, we keep a single narrow backbone network $\fF$ but each additional broad view $x^k_{b}$ has its own backbone $\fB^k$. The projector and predictor are shared across modalities. 
	Given this, all regression losses are simply aggregated over all pairs composed by the narrow view $x_n$ and the different broad views $\{x^k_{b}\}_k$:
	\begin{align}
		\mathcal{L}(x) = \sum_{k=1}^K \mathcal{L}^k_{n\rightarrow b}(x) + \mathcal{L}^k_{\bton}(x).
	\end{align}
	When using different modalities, the risk for the broad network to only focus on the narrow view is reduced due to the modality gap between the two views. 
	Furthermore, when using audio, syncing helps slightly as previously observed in visual-audio work~\cite{korbar2018cooperative}.
	\added{We analyse further the audio-visual syncing strategies in Appendix Section~\ref{tab:sync_apendix}}.

\noindent
\textbf{Final loss.} Given a large set of videos $\{x^i\}_{i=1}^N$, we train our model to minimize:
\begin{align}
	\min_{\substack{\fF, \fB^k, \pF,\qF}} \sum_{i=1}^N \mathcal{L}(x^i).
\end{align}
Next, we provide more details on the specific components that are used when \algo is applied in the unimodal and multimodal settings; as well as how the narrow and broad views are constructed in each case.
	
\subsection{Broad views from visual and audio modalities}
\label{subsect:multimodal}

In our framework, we regress the representation of a broad backbone which sees a larger context of the video. The \targetview is meant to provide information about the full video clip including more temporal context, in order to supervise the narrow backbone $\fF$. 
As the different views are processed by different backbones, we can apply a different set of pre-processing and augmentation functions to any of the views. 
In this section, we first describe the set of transformations that we use when training with visual inputs alone, and then when training with both visual and audio inputs.

\noindent
\textbf{Visual modalities.}
When sampling the \targetview from the visual modalities, we aim to cover a large temporal context, the full clip. Accessing more temporal context typically means increasing the number of frames, and thus introducing extra computational complexity. To avoid this overhead, we decrease the spatial resolution of the \targetview in order to keep the number of pixels constant. In Section~\ref{sec:results} we show the effectiveness of trading temporal context for spatial resolution in the \targetview. 
By keeping the computational cost fixed, we ensure that our method is computationally competitive with alternative self-supervised approaches.

Additionally to the temporal sampling, the set of transformations we consider for use on the narrow and broad views  are motivated from two complementary perspectives. 
First, similarly to the use of augmentations in a wide number of machine learning approaches, and in particular in contrastive and regression-based self-supervised learning approaches, we also employ such stochastic transformations to enforce invariance or equivariance constraints on the learned representations.
In contrast to the use of augmentations in these self-supervised frameworks however, we emphasize that we do not impose that the set of transformations $\mathcal{T}_n$  used on the \sourceview be the same as the set of transformations $\mathcal{T}_b$  used on the \targetviews.
To explore this, we employ a recently introduced augmentation procedure relying on random convolutions~\cite{xu2020robust}, by which we augment only the \targetview. \added{Details about the random convolutions can be found in the Appendix ~\ref{subsec:randomconv}.}

Second, we can design the transformations $\mathcal{T}_b$ used for the \targetview to extract specific features from the input modality, sought to enrich the learned representations $\fF(\xF)$ with a  certain type of information. For example, we can use optical flow as substitute of RGB in the \targetview, which is reminiscent of~\cite{Stroud18}, where the flow network is used to teach the RGB network. Optical flow from sequential images can provide supervision to emphasize motion in the learned representations extracted from the source, which has shown to be important for predicting actions~\cite{simonyan2014,walker2016uncertain,Han20}.
Optical flow can be extracted using an off-the-shelf unsupervised flow extraction algorithm. 
As flow is computed once for the full dataset, its computational overhead is negligible compared to training time.

\noindent \textbf{Audio modalities.}
\label{subsect:audiovisual}
Our framework can leverage audio as supervisory signal in the \targetview. 
We can either use a single audio broad view or combine a visual broad view and an audio broad view for stronger self-supervision. 
Audio is a strong supervisory signal, and has been extensively used for self-supervision in videos
as it strongly correlates with the visual content, while being easier to process computationally. 
As pre-processing, we extract spectrograms from consecutive short-time windows on the waveform using Fourier transforms. This approach has been shown to be very effective in obtaining state-of-the-art performance on supervised \cite{ford2019deep, kong2020panns} and unsupervised \cite{jansen2018unsupervised, jansen2020coincidence, alayrac2020self} approaches. 
For this reason, we encode the audio using a log-mel spectrogram representation as $\xB \in \mathbb{R}^{T_s \times D}$ where $T_s$ is the number of spectrogram frames and $D$ denotes the number of features.
Similar to the unimodal setting, we experiment with enlarging the temporal window for the extraction of the audio view, compared with the temporal window of the narrow video view, seeking to increase the amount of context information present in the supervisory signal. Finally, as explained in the previous section, we make sure that the visual \sourceview and the audio \targetview are in sync at their starting point.


\section{Experiments}
\label{sec:results}\noindent
In this section, we evaluate \algo and compare its performance against relevant state-of-the-art methods trained on similar data and modalities. Further details about the experimental setting can be found in the Appendix. 
\subsection{Experimental setting}

\noindent
\textbf{Video-only experiments.}
In the video-only setting, unless stated otherwise we conduct our experiments on the Kinetics-600 dataset~\cite{carreira2018short}.
The dataset has 600 action classes and contains $447$k videos at the time of experimentation, $362$k in the train set. We also train on the Kinetics-400~\cite{kay2017kinetics} dataset for comparison with the state-of-the-art. 

\noindent
\textbf{Audio-video experiments.}
In the crossmodal training setting, we use the AudioSet~\cite{gemmeke2017audio} as pre-training dataset.
The dataset has 527 action classes and contains $1.9$M videos in the training set at the time of experimentation. 

\noindent
\textbf{Architectures.}
For spatiotemporal volumes such as the sequences of RGB or flow frames, unless specified otherwise, we use the TSM-ResNet50 (TSM-50)~\cite{lin2019tsm} architecture for the narrow backbone. 
For the broad visual backbone we always use a TSM-50 backbone.
Video inputs are sampled at $12.5$ frames per second (FPS), except when using the R3D architecture which we train sampling videos at $6.25$ FPS.
Unless stated otherwise, we train the narrow backbone on inputs of 16 frames (1.3 seconds) at resolution $224\times224$, and the broad backbone on inputs of $64$ frames at $6.25$ FPS (10s) at resolution $112\times112$.
To see how our method scales to different and bigger architectures, we also experiment with different backbones for the narrow network with the R(2+1)D architecture~\cite{tran2018closer}, R3D architecture (as described in \cite{feichtenhofer2021large}) and TSM with twice the number of channels in each layer (TSM-50x2).
We use these networks only for the \sourceview.
For the broad backbone processing log-mel spectrograms, we use ResNet-50~\cite{he16resnet}.
\added{All models are trained using a two-layer MLP for the projector and predictor heads with a hidden layer of dimension $4096$. We use batch normalization after each hidden layer. In the projector heads, we use batch normalisation after the last layer.} 
We use $128$ as the output dimension of projectors and predictors. 

\noindent
\textbf{Feature extraction.} 
For flow extraction, we use the TV-L1~\cite{zach2007duality} algorithm.
We use 80 bins for extracting log-mel spectrograms.

\noindent
\textbf{Augmentations.}
We sample and augment all the visual views independently. 
For any narrow view, we uniformly sample a temporal offset between $0$ and $T - \tau_n$, where $T$ is the duration of the video clip and $\tau_n$ denotes the length of the narrow view. We extract the view starting at this offset.
For the \targetview, we randomly sample the offset between $0$ and $T$. We pad any broad view of insufficient length with a clip extracted from the start of the video sample (\ie looping over the sequence).
For all visual modalities (including the flow), we use random cropping and horizontal flipping. 
For the RGB views, we additionally employ Gaussian blurring as well as scale and color jittering.
We also explore the use of random convolutions as an augmentation procedure. 
Following~\cite{Lee2020Network}, we use He initialization~\cite{he2015delving} for the weights and fixed zero bias, sampling the size of the kernel uniformly across odd values ranging from 1 to 11. 
For audio, we use the same starting point as the narrow view, but extend it for a longer time window. 
If necessary, similarly to the RGB case, we pad the broad audio view with audio extracted from the start of the audio clip.

\noindent
\textbf{Self-supervised training details.}
We discard labels at training time, and only use them for downstream evaluation.
\added{
We employ a batch size of 512 and train for 300k steps, setting the initial learning rate to $4.8$ for models without audio and $1.0$ for models with audio.
We train all models using LARS~\cite{you2018imagenet}.
We use 5000 warm up steps and cosine learning rate schedule~\cite{loshchilov2016sgdr}.}
Following BYOL~\cite{grill2020bootstrap}, we multiply the learning rate for predictor $\qF$ by 10.
For batch norm layers, we use a decay rate of 0.9 and epsilon of 1e-5. 
We use weight decay of $0.01$ except for the R(2+1d)-18 architecture where we use $10^{-7}$.
\added{As in \cite{feichtenhofer2021large}, we do not apply LARS and weight decay to the biases and batch norm parameters.}
\subsection{Downstream tasks}

We use two standard settings to evaluate the quality of the learned visual representations from the narrow backbone $\fF$:
in the \emph{linear} setting, we train a linear layer over frozen features extracted by $\fF$;
in the \emph{fine-tuning} setting, we train $\fF$ and the classifier head end-to-end.
\added{Unless stated otherwise, we use $32$ frames for video evaluation, to be comparable to previous work.}
We evaluate video representations using the HMDB51 dataset~\cite{kuehne2011hmdb}, the UCF101 dataset~\cite{soomro2012} and the Kinetics-600~\cite{carreira2017quovadis} validation set.
The HMBD51 dataset contains 5K videos, corresponding to 51 classes. 
The UCF101 dataset contains 13K videos, corresponding to 101 classes.
The Kinetics-600 validation set contains $28$k videos.
We also evaluate the learned audio representations from the corresponding broad backbone, $\fB$, on both the test set of the AudioSet dataset (20K samples, 527 classes) as well as the smaller ESC-50 dataset~\cite{piczak2015dataset} (2K samples, 50 classes). 
Following standard procedure, we report top-1 accuracy for all datasets except for Audioset where we report the mean average precision~\cite{jansen2020coincidence}.
For the datasets that have official splits (3 for UCF101/HMDB51 and 5 for ESC-50), we follow the standard procedure where split\#1 serves as the validation set and the average accuracy over all splits is then reported.

\noindent
\textbf{Linear setting.}
For HMDB51, UCF101 and ESC-50, we extract representations from 10 epochs worth of augmented samples using the learned narrow backbone, and we train a linear SVM using scikit-learn~\cite{pedregosa2011scikit} on these frozen features. 
\added{For Kinetics-600 and AudioSet which are larger, we instead train the linear classifier using the LARS~\cite{you2018imagenet} optimiser for Kinetics-600 and the Adam optimizer~\cite{kingma15adam} for AudioSet.}
In all cases, we use the same augmentations as during unsupervised pre-training except for gaussian blur.
At test time, we average the prediction over 30 clips (10 temporal clips each with 3 spatial crops) as done in~\cite{qian2020spatiotemporal}.
For AudioSet, we follow~\cite{jansen2020coincidence} and use a fully-connected classifier, with one hidden layer of 512 units, in place of the linear classifier.  

\noindent
\textbf{Fine-tuning setting.} In this setting, we add a single, randomly initialized, linear layer at the output of the narrow backbone. 
We initialize the narrow backbone's weights with those learned using \algo, and we fine-tune this architecture end-to-end.
Following previous work, we perform this evaluation on the HMDB51 and UCF101 datasets. 
At test time, we average the prediction over 30 clips (10 temporal clips each with 3 spatial crops).

	\begin{table}[t]
		\centering
		\caption{\small {\bf Sync views.} Effect of syncing the narrow and broad views.
        }
        \vspace{0.5em}
		\resizebox{\linewidth}{!}{
		\begin{tabular}{ccc|ccc} \toprule
			Dataset & Sync  & $M_b$ & HMDB51 &UCF101& K600 \\ 
			\midrule
			\midrule
			K600 & \xmark & RGB+RC & \bf 65.2 & \bf 91.7 & \bf 69.1  \\
			K600 & \cmark & RGB+RC & 64.3 & 87.5 & 60.5 \\

			\bottomrule
		\end{tabular}
	}
		\label{tab:sync}
	\end{table}
 
	\begin{table}[t]
		\centering
		\caption{\small {\bf Importance of the broad view.} We evaluate the impact of the temporal extent of the narrow ($\tau_n$) and broad ($\tau_b$) views.
		$M_b$ is the modality used in the broad view.
		RC stands for random convolutions. 
		K600 stands for Kinetics-600 and AS for AudioSet.
        }

        \vspace{0.5em}
		\resizebox{\linewidth}{!}{
		\begin{tabular}{cccc|ccc} \toprule
			Dataset & $M_b$ & $\tau_n$ & $\tau_b$ & HMDB51 &UCF101 & K600  \\ 
			\midrule
			\midrule
			 K600 & RGB+RC& 10s  & 10s & 58.0 & 79.6  & 47.1 \\
			 K600 & RGB+RC & 1.3s & 1.3s & 59.8  &89.5 & 67.1 \\
			 K600 & RGB+RC & 1.3s & 5s & 63.3&90.5  &66.7 \\
			 K600  & RGB+RC & 1.3s & 10s & \textbf{65.2} & \textbf{91.7} &\textbf{69.1} \\
			\midrule
			\midrule
			 AS & Audio & 1.3s & 1.3s &  68.1 &92.3 & 68.8  \\
			 AS& Audio & 1.3s & 5s & \bf 68.2 & \bf 92.7& 69.7 \\
			 AS& Audio & 1.3s & 10s & 67.1 & 92.5  & \bf 70.0  \\
			\bottomrule
		\end{tabular}
	}
		\label{tab:ablation_res}
	\end{table}
	
	\begin{table}[t]
		\centering
		\caption{\small {\bf Visual transformation for the broad view.}  
		We compare various augmentations for the visual input of the broad view, when pre-training on Kinetics-600.
		We use $\tau_n = 1.3s$ (narrow extent) and $\tau_b=10s$ (broad extent).
		RC stands for random convolutions. 
        }
        \vspace{0.5em}
		\resizebox{0.7\linewidth}{!}{
		\begin{tabular}{c|ccc} \toprule
			$M_b$ & HMDB51 &UCF101 & K600 \\ 
			\midrule
			\midrule
			 RGB & 63.2 & 90.5 & \bf 69.3 \\
			 RGB+RC &65.2 & 91.7 & 69.1  \\
			 Flow & \textbf{66.4} & \textbf{92.1} & 67.0   \\
			\bottomrule
		\end{tabular}
	}
		\label{tab:ablation_modality}
	\end{table}

\subsection{Ablation study}

In this section, we study the effect of the different components of \algo on the performance of the narrow backbone $\fF$.
Specifically, we study five main elements:
\textbf{(i)} the effect of temporally syncing the \sourceview and the \targetview, \textbf{(ii)} the effect of the temporal extents of the narrow and  \targetviews, \textbf{(iii)} the improvements brought by different choices of transformations for the visual modality, \added{\textbf{(iv)} the improvements resulting from using multiple broad views of the same modality} and  \textbf{(v)} the effect of using a separate predictor and projector for each pair of views.
\added{
We conduct this analysis using the HMDB51, UCF101 and K600 benchmarks in the linear setting.
Further discussion on syncing audio and video, using \algo to learn image models, evaluating the broad backbone or training with broad RGB and flow views at the same time can be found in the Appendix.}

\noindent
\textbf{Syncing views.}
In Table~\ref{tab:sync}, we study the effect of having the same temporal starting point for the narrow and the broad view. 
\added{As expected, when using a broad visual modality, syncing significantly decreases performance.} 
We hypothesise that when both views are in sync, the broad network can simply focus its prediction only on the narrow view since the relative position of the views is deterministic hence making the self-supervised task easier as explained in Section~\ref{subsect:framework}.
 
\noindent
\textbf{Importance of the broad view.}
We study the effect of the temporal extent of the narrow and broad views in the RGB-only setting (using random convolutions RGB+RC for the broad view) and the multimodal setting (using audio spectrogram for the broad view).
We report results in Table~\ref{tab:ablation_res}.
First, in the unimodal setting, we find that for a \sourceview extent $\tau_n$ of $1.3s$, performance improves significantly across the two downstream tasks as we increase the duration of the broad view $\tau_b$ from $1.3s$ to $10s$, \added{(\eg from 59.8 to 65.2 on HMDB51)}. 
This empirically supports our intuition that broader views can provide better supervision.
Second, we find that using temporally large views of $10s$ for both the \sourceview and the \targetview degrades performance, as the task becomes significantly easier and we are unlikely to get rich embeddings.
\added{In the multimodal setting, we find that increasing the context from $1.3s$ to $5s$ brings an improvement to UCF101 and specially Kinetics-600, although it is smaller than in the visual setting. Since extending to $10s$ does not result in additional improvements, we use $5s$ for the audio broad view. }

	\begin{table}[t]
		\centering
		\caption{\small {\bf Number of broad views.} Effect of adding multiple broad views of the same modality (RGB+RC).
        }
        \vspace{0.5em}
		\resizebox{\linewidth}{!}{
		\begin{tabular}{cc|ccc} \toprule
			Dataset & Number of views & HMDB51 &UCF101& K600 \\ 
			\midrule
			\midrule
			K600 & 1 & 65.2 & 91.7 & 69.1  \\
			K600 & 2 & \bf 65.4  & 91.8 & 70.4 \\
			K600 & 3 & \bf 65.4 & \bf 92.6 & \bf 70.8  \\

			\bottomrule
		\end{tabular}
	}
		\label{tab:num_views}
	\end{table}

	\begin{table}[t]
		\centering
		\caption{\small {\bf Weight sharing.} 
		We explore the effect of sharing weights across different components of the models.
		Models are trained on the Kinetics-600 dataset using RGB visual input in the broad view.
        }
        \vspace{0.5em}
		\resizebox{\linewidth}{!}{
		\begin{tabular}{ccc|ccc} \toprule
			Shared  & Shared  & Shared  & \multirow{2}{*}{HMDB51} & \multirow{2}{*}{UCF101} & \multirow{2}{*}{K600} \\ 
			Backbone & Projector & Predictor &  &  \\ 

			\midrule
			\midrule
			\xmark & \xmark & \xmark &  61.3 & 89.9 & 67.7 \\
			\xmark & \cmark & \xmark & 61.7  & 90.2 & 68.6  \\
			\xmark & \cmark & \cmark & \bf 63.2  & \bf 90.5 & \bf 69.3  \\
			\hline
			\spv{\cmark} & \spv{\cmark} & \spv{\xmark} &  \spv{\bf 63.7} & \spv{\bf 91.4} & \spv{\bf 69.4}  \\
			\spv{\cmark} & \spv{\cmark} & \spv{\cmark} &  \spv{62.7} & \spv{91.2} & \spv{69.3} \\

			\bottomrule
		\end{tabular}
	}
		\label{tab:ablation_shared}
	\end{table}

\vskip 3mm	
\noindent

\textbf{Visual transformation for the broad view.} 
In Table~\ref{tab:ablation_modality}, we investigate the effect of using different visual inputs in the \targetview. 
First, we see that using Random Convolutions (RC)~\cite{xu2020robust} on the RGB frames significantly improves performance, compared to using standard RGB frames.
\algo enables the use of such an aggressive augmentation since it has a dedicated backbone for that view.
Moreover, only using this augmentation on the \targetview ensures that the backbone trained on the \sourceview does not suffer from shift in distribution of intensities~\cite{xu2020robust}.
\added{Furthermore, using optical flow for the \targetview leads to further improvement in HMDB51 and UCF101 when compared to using RC augmentation.}
This demonstrates a surprisingly high effectiveness of leveraging hand-designed features, probably because this allows important factors -- here motion and segmentation information -- to be included in the representation. Finally, we see how the performance in K600 decreases when flow is used in the broad view. We attribute this to the semantic nature of the dataset, which requires richer semantic information than the one provided by flow.

	\begin{table*}[t]
		\centering
		\caption{\small {\bf Comparison with the state-of-the-art.} We report performance in the linear and fine-tuning (FT) settings, on 3 vision benchmarks: UCF101, HMDB51, Kinetics-600 (K600); and 2 audio benchmarks: ESC-50 and AudioSet (AS). K400 is Kinetics-400, YT8M is Youtube-8M~\cite{abu2016youtube}, IG65M is Instagram-65M~\cite{ghadiyaram2019large}, IN is ImageNet~\cite{imagenet}. We specify dataset sizes in years.
		We denote the modalities $\mathcal{M}$ used for training by: V for RGB, F for flow and A for audio. \textbf{All models use only RGB for the visual downstream tasks.}
		}
		\vspace{0.5em}
		\tablestyle{2pt}{1.05}
		\resizebox{\linewidth}{!}{
		\begin{tabular}{lrcccccccc|cc} \toprule
			& & & && \multicolumn{2}{c}{UCF101} & \multicolumn{2}{c}{HMDB51} & K600 & ESC-50 & AS  \\
			\cmidrule(lr){6-7} \cmidrule(lr){8-9}
			Method & Backbone (\#params)&  Dataset & Years & $\mathcal{M}$ & Linear & FT & Linear & FT & Linear & Linear  & MLP  \\ \midrule
			CoCLR~\cite{Han20}  & S3D (9.1M) &  K400 & 0.07 & VF & 74.5  & 87.9 & 46.1 & 54.6 &   & / & /  \\
			CVRL~\cite{qian2020spatiotemporal} & R3D50 (31.8M) &  K600 & 0.1 & V & 90.6 & 93.4  & 59.7 & 68.0 & 70.4 & / & /  \\
			
			$\rho$BYOL~\cite{feichtenhofer2021large} & R3D50 (31.8M) &  K400 & 0.07 & V &  & 95.5  &  & 73.6 &  & / & /  \\
			$\rho$BYOL~\cite{feichtenhofer2021large} & S3D (9.1M) &  K400 & 0.07 & V &  & \bf 96.3  &  & 75.0 &  & / & /  \\

			\midrule
			\algo:V$\leftrightarrow$V$\times 3$ (ours) & R3D50 (31.8M) &  K400 & 0.07 & V & 90.6  &94.7 &64.4  &71.9 & 66.7  & / & /  \\
		    \algo:V$\leftrightarrow$F$\times 3$  (ours) & R3D50 (31.8M) &  K400 & 0.07 & VF & 92.5  &95.1 & 68.3 &74.6 &  66.9 & / & /  \\

			\algo:V$\leftrightarrow$V$\times 3$ (ours) & TSM-50 (23.5M) &  K600 & 0.1 & V   &  92.6 &94.7 &65.4  &74.6 & \bf 70.8  & / & /  \\
			\algo:V$\leftrightarrow$F$\times 3$ (ours) & TSM-50 (23.5M) &  K600 & 0.1 & VF  & 92.9  &95.2 & 66.8 &74.9 & 68.1  & / & /  \\

			\algo:V$\leftrightarrow$V$\times 3$ (ours) & R3D50 (31.8M) &  K600 & 0.1 & V&  92.4 &95.0 & 66.7 & 74.2& 70.0  & / & /  \\
			\algo:V$\leftrightarrow$F$\times 3$  (ours) & R3D50 (31.8M) &  K600 & 0.1 & VF& \bf 93.1  &  95.4 & \bf 69.5 &\bf 76.0 &   69.2& / & /  \\

			\midrule
			\midrule
			GDT~\cite{mandela2020datatrans}
			& R(2+1)D-18 (33.3M) & AS & 1 & VA & & 92.5 & & 66.1  &  & 88.5&  \\
			MMV~\cite{alayrac2020self} & R(2+1)D-18 (33.3M) & AS & 1 & VA &
			83.9 & 91.5 & 60.0 & 70.1 & 55.5  & 85.6 & 29.7  \\  %
			XDC~\cite{alwassel2019self} & R(2+1)D-18 (33.3M) & AS & 1 & VA & & 93.0 & & 63.7 & & 84.8  & \\

			XDC~\cite{alwassel2019self} & R(2+1)D-18 (33.3M) &  IG65M & 21 & VA & & 95.5 & & 68.9 &  & 85.4 & \\
			ELo~\cite{piergiovanni2020evolving}
			& R(2+1)D-50 (46.9M) & YT8M & 13 & VFA & & 93.8 & 64.5 & 67.4 &  &  & \\
			AVID~\cite{morgado20avid}
			& R(2+1)D-50 (46.9M) & AS & 1 & VA &
			& 91.5 & & 64.7  &  & 89.2 &  \\

			\midrule
		    \algo:V$\leftrightarrow$A (ours) & R(2+1)D-18 (33.3M)  &AS & 1 & VA & 90.0 & 93.6 & 63.6 & 70.8& 64.1 & 91.9 & \bf 36.5 \\
			\algo:V$\leftrightarrow$A (ours) & TSM-50 (23.5M)  &AS & 1 & VA &  93.0 & 95.6& 69.4 &76.5 &  \bf 70.6 & \bf 93.0 & 36.4   \\
			\algo:V$\leftrightarrow$FA (ours) & TSM-50 (23.5M)  &AS & 1 & VFA &  92.9 &95.6 & 69.2 & 78.0 & 69.3  & 92.0 & 35.2  \\
			\algo:V$\leftrightarrow$FA (ours) & R(2+1)D-50 (46.9M)  &  AS & 1 & VFA & 92.5 & 95.0 &66.6 & 73.5 & 68.7 & 92.3 & 35.5 \\

			\algo:V$\leftrightarrow$FA (ours) & TSM-50x2 (93.9M)  &  AS & 1 & VFA & \bf 93.2  & \bf 96.9& \bf 69.9 & \bf 79.4 & 70.3  & 91.7  & 36.2  \\
			\midrule
			\spv{S3D~\cite{xie2018rethinking}} &  \spv{S3D (9.1M)}  & \spv{IN+K400}& \spv{0.07} & & &\spv{96.8} & &\spv{75.9}& & \\
			\spv{R(2+1)D-18~\cite{alwassel2019self}} &  \spv{R(2+1)D-18 (33.3M)}  & \spv{K400}& \spv{0.07} & & \spv{} & \spv{94.2}& &\spv{65.1} & &&\\
			\spv{R(2+1)D-50~\cite{piergiovanni2020evolving}} &  \spv{R(2+1)D-50 (46.9M)}  & \spv{K400}& \spv{0.07} &
			\spv{} & & \ & \spv{71.5} &   & & &   \\
			\spv{SMART~\cite{gowda2020smart} \footnotemark[1]} & \spv{TSN-Inception-V3 (24M)}  &  \spv{K400}& \spv{0.07} &
			\spv{} & & \spv{98.6} &  & \spv{84.3}  & &   &    \\
			\spv{TimeSformer-L~\cite{bertasius2021space}} & \spv{TimeSformer-L (121.4M)}  & \spv{IN+K600} & \spv{0.1} &
			\spv{} & &  & &  &\spv{82.4} \footnotemark[2] &  &   \\
			\spv{AST~\cite{gong2021ast}} & \spv{ViT-B (87.0M)}  &\spv{AS} & \spv{1} &
			\spv{} & &  &  &  & & \spv{95.6} \footnotemark[2] & \spv{48.5} \footnotemark[2]   \\

			\bottomrule
		\end{tabular}
	}
		\label{tab:sota}
		\vspace*{-0.2cm}
	\end{table*}

\noindent \textbf{Number of broad views.}
\added{
In Table~\ref{tab:num_views}, we study the impact of having more than one broad view of the same modality. Adding additional broad views results in improved performance on UCF-101 and Kinetics. We believe that using multiple views serves as augmentation for \algo, which regresses to the average of multiple projection and this is likely to be more representative of the full video.}

\noindent \textbf{Sharing weights.}
In Table~\ref{tab:ablation_shared}, we study the effect of sharing weights on different parts of the overall system. Specifically, we focus on studying \algo trained with RGB broad view (without random convolution), a setting in which we can study sharing all parts of the model. However, when using broad views with different modalities such as audio or flow, using a single backbone is not desirable as we cannot use domain specific backbones which are useful for good performance. The results show that when sharing more parameters, the model improves its performance on various benchmarks. However, we observe that it is important to at least have one module with separate weights across views. Furthermore, we find specially important to share the projector layer, which seems to lead to larger performance difference. For simplicity and flexibility, in the final version of \algo we use separate backbones for the narrow and broad views and a single predictor and projector.

\subsection{Comparison with the state-of-the-art}
\footnotetext[1]{This model utilises a lightweight network for frame selection and a larger model (Inception-V3) for classification.}
\footnotetext[2]{This model has been train end-to-end with a supervised loss.}

We compare \algo against the state-of-the-art for self-supervised video representation learning in Table~\ref{tab:sota}.
Note that when evaluating in visual tasks, we only use the RGB modality to be comparable to previous work.

\noindent
\textbf{Visual only on Kinetics.}
\added{
In the setting where we use only the video modality combined with three broad views using random convolutions, we find that \algo outperforms the CVRL approach~\cite{qian2020spatiotemporal} on UCF101 and HMDB both linear and fine-tuning under similar conditions (R3D and Kinetics-600). Furthermore, when integrating the flow modality in the broad view and using similar backbone (R3D) and dataset (K400), \algo performs competitively with $\rho$-BYOL~\cite{feichtenhofer2021large} ($-0.4\%$ in UCF101 and $+1\%$ in HMDB51).  
Note that differently from \cite{feichtenhofer2021large}, our method does not require an EMA network, which introduces additional computational requirements. Moreover, we observe that the TSM architecture does perform comparably to R3D in UCF101 and Kinetics-600 although having less parameters overall. 
Finally, we observe that using the flow modality increases the performance for most of the settings.
}

\noindent
\textbf{Multimodal on AudioSet.}
We also compare our approach in the multimodal (visual and audio modalities) setting by training \algo on AudioSet.
In that setting, we train for $620$k steps instead of $300$k, as AudioSet is significantly larger than Kinetics-600. We increase the number of input frames of the narrow network from $16$ to $32$ frames (at 12.5FPS). We use $\tau_b=5s$ for the audio broad view.
\added{
We make four important observations.
\textbf{(i)} Under this setting, \algo outperforms all state-of-the-art methods when using the same pretraining data and backbone. 
In particular, when using R(2+1d)-18 we outperform the current state-of-the-art XDC~\cite{alwassel2019self} and MMV~\cite{alayrac2020self} which uses the same architecture and training dataset.
\textbf{(ii)} Interestingly, we observe that using two broad views coming from two different modalities, audio and flow, benefits performance on HMDB51 (+1.5\%) but performs similarly in UCF101. 
\textbf{(iii)} \algo benefits from using larger visual backbones. 
When using the larger backbone TSM-50x2 (93.9M parameters), \algo establishes a new state-of-the-art on HMDB51 finetuning with $79.4$ and UCF101 finetuning with $96.9$. This performance improvement brings the self-supervised results one step closer to the best supervised models, which use additional techniques such as frame filtering and larger backbones. Furthermore, \algo is competitive with baselines using a single convolutional architectures and RGB input~\cite{alwassel2019self,xie2018rethinking}, which is comparable with our evaluation setting. 
\textbf{(iv)} When evaluating the performance of the broad audio network we also significantly outperform previous state-of-the-art on two challenging benchmarks, ESC-50 and Audioset.
Notably, we significantly improve the performance in AudioSet, the hardest of the audio tasks. 
}

\section{Conclusion}
\label{sec:conclusions}
In this paper, we introduced \algo, a self-supervised learning framework for video. Our method learns representations by supervising a temporally narrow view with a general broad view, which can be either computed from RGB, flow or audio. 
Our model achieves state-of-the-art performance when trained on datasets such as Kinetics or AudioSet. 
Notably, when trained with a larger backbone, \algo is competitive with supervised pre-trained techniques.

\paragraph{Acknowledgments.}
The authors would like to thank  Antoine Miech, Bilal Piot, Evan Shelhamer and Sander Dieleman for fruitful discussions.

{\small
\bibliographystyle{ieee_fullname}
\bibliography{main}

\begin{thebibliography}{10}\itemsep=-1pt

\bibitem{abu2016youtube}
Sami Abu-El-Haija, Nisarg Kothari, Joonseok Lee, Paul Natsev, George Toderici,
  Balakrishnan Varadarajan, and Sudheendra Vijayanarasimhan.
\newblock Youtube-8m: A large-scale video classification benchmark.
\newblock {\em arXiv preprint arXiv:1609.08675}, 2016.

\bibitem{alayrac16unsupervised}
Jean-Baptiste Alayrac, Piotr Bojanowski, Nishant Agrawal, Ivan Laptev, Josef
  Sivic, and Simon Lacoste-Julien.
\newblock Unsupervised learning from narrated instruction videos.
\newblock In {\em CVPR}, 2016.

\bibitem{alayrac2020self}
Jean-Baptiste Alayrac, Adri{\`a} Recasens, Rosalia Schneider, Relja
  Arandjelovi{\'c}, Jason Ramapuram, Jeffrey De~Fauw, Lucas Smaira, Sander
  Dieleman, and Andrew Zisserman.
\newblock Self-supervised multimodal versatile networks.
\newblock In {\em NeurIPS}, 2020.

\bibitem{alwassel2019self}
Humam Alwassel, Dhruv Mahajan, Lorenzo Torresani, Bernard Ghanem, and Du Tran.
\newblock Self-supervised learning by cross-modal audio-video clustering.
\newblock In {\em NeurIPS}, 2020.

\bibitem{arandjelovic17look}
Relja Arandjelovi\'{c} and Andrew Zisserman.
\newblock Look, listen and learn.
\newblock In {\em ICCV}, 2017.

\bibitem{arandjelovic2018objects}
Relja Arandjelovi\'{c} and Andrew Zisserman.
\newblock Objects that sound.
\newblock In {\em ECCV}, 2018.

\bibitem{asano2020selflabelling}
Yuki~Markus Asano, Christian Rupprecht, and Andrea Vedaldi.
\newblock Self-labelling via simultaneous clustering and representation
  learning.
\newblock In {\em ICLR}, 2020.

\bibitem{AMDIM}
Philip Bachman, R~Devon Hjelm, and William Buchwalter.
\newblock Learning representations by maximizing mutual information across
  views.
\newblock In {\em Neural Information Processing Systems}, 2019.

\bibitem{bachman2019learning}
Philip Bachman, R~Devon Hjelm, and William Buchwalter.
\newblock Learning representations by maximizing mutual information across
  views.
\newblock In {\em NeurIPS}, 2019.

\bibitem{bautista2016cliquecnn}
Miguel~A. Bautista, Artsiom Sanakoyeu, Ekaterina Sutter, and Björn Ommer.
\newblock Cliquecnn: Deep unsupervised exemplar learning.
\newblock In {\em NeurIPS}, 2016.

\bibitem{bertasius2021space}
Gedas Bertasius, Heng Wang, and Lorenzo Torresani.
\newblock Is space-time attention all you need for video understanding?
\newblock {\em arXiv preprint arXiv:2102.05095}, 2021.

\bibitem{caron2018DeepCF}
Mathilde Caron, Piotr Bojanowski, Armand Joulin, and Matthijs Douze.
\newblock Deep clustering for unsupervised learning of visual features.
\newblock In {\em ECCV}, 2018.

\bibitem{caron2020unsupervised}
Mathilde Caron, Ishan Misra, Julien Mairal, Priya Goyal, Piotr Bojanowski, and
  Armand Joulin.
\newblock Unsupervised learning of visual features by contrasting cluster
  assignments.
\newblock In {\em NeurIPS}, 2020.

\bibitem{carreira2018short}
Joao Carreira, Eric Noland, Andras Banki-Horvath, Chloe Hillier, and Andrew
  Zisserman.
\newblock A short note about kinetics-600.
\newblock {\em arXiv preprint arXiv:1808.01340}, 2018.

\bibitem{carreira2017quovadis}
Joao Carreira and Andrew Zisserman.
\newblock Quo vadis, action recognition? {A} new model and the {Kinetics}
  dataset.
\newblock In {\em CVPR}, 2017.

\bibitem{chen2020simple}
Ting Chen, Simon Kornblith, Mohammad Norouzi, and Geoffrey Hinton.
\newblock A simple framework for contrastive learning of visual
  representations.
\newblock In {\em ICML}, 2020.

\bibitem{hjelm2020representation}
R Devon et~al.
\newblock Representation learning with video deep infomax.
\newblock {\em arXiv preprint arXiv:2007.13278}, 2020.

\bibitem{doersch2015unsupervised}
Carl Doersch, Abhinav Gupta, and Alexei~A Efros.
\newblock Unsupervised visual representation learning by context prediction.
\newblock In {\em ICCV}, 2015.

\bibitem{doersch2017multi}
Carl Doersch and Andrew Zisserman.
\newblock Multi-task self-supervised visual learning.
\newblock In {\em ICCV}, 2017.

\bibitem{dosovitskiy2014discriminative}
Alexey Dosovitskiy, Jost~Tobias Springenberg, Martin Riedmiller, and Thomas
  Brox.
\newblock Discriminative unsupervised feature learning with convolutional
  neural networks.
\newblock In {\em NIPS}, 2014.

\bibitem{faghri2017vse}
Fartash Faghri, David~J Fleet, Jamie~Ryan Kiros, and Sanja Fidler.
\newblock Vse++: Improving visual-semantic embeddings with hard negatives.
\newblock In {\em BMVC}, 2017.

\bibitem{feichtenhofer2021large}
Christoph Feichtenhofer, Haoqi Fan, Bo Xiong, Ross Girshick, and Kaiming He.
\newblock A large-scale study on unsupervised spatiotemporal representation
  learning.
\newblock In {\em Proceedings of the IEEE/CVF Conference on Computer Vision and
  Pattern Recognition}, pages 3299--3309, 2021.

\bibitem{ford2019deep}
Logan Ford, Hao Tang, Fran{\c{c}}ois Grondin, and James~R Glass.
\newblock A deep residual network for large-scale acoustic scene analysis.
\newblock In {\em InterSpeech}, 2019.

\bibitem{gemmeke2017audio}
Jort~F Gemmeke, Daniel~PW Ellis, Dylan Freedman, Aren Jansen, Wade Lawrence,
  R~Channing Moore, Manoj Plakal, and Marvin Ritter.
\newblock Audio set: An ontology and human-labeled dataset for audio events.
\newblock In {\em ICASSP}, 2017.

\bibitem{ghadiyaram2019large}
Deepti Ghadiyaram, Du Tran, and Dhruv Mahajan.
\newblock Large-scale weakly-supervised pre-training for video action
  recognition.
\newblock In {\em CVPR}, 2019.

\bibitem{gidaris2020learning}
Spyros Gidaris, Andrei Bursuc, Nikos Komodakis, Patrick Pérez, and Matthieu
  Cord.
\newblock Learning representations by predicting bags of visual words.
\newblock In {\em CVPR}, 2020.

\bibitem{gidaris2018unsupervised}
Spyros Gidaris, Praveer Singh, and Nikos Komodakis.
\newblock Unsupervised representation learning by predicting image rotations.
\newblock In {\em ICLR}, 2018.

\bibitem{gong2021ast}
Yuan Gong, Yu-An Chung, and James Glass.
\newblock Ast: Audio spectrogram transformer.
\newblock {\em arXiv preprint arXiv:2104.01778}, 2021.

\bibitem{gowda2020smart}
Shreyank~N Gowda, Marcus Rohrbach, and Laura Sevilla-Lara.
\newblock Smart frame selection for action recognition.
\newblock {\em arXiv preprint arXiv:2012.10671}, 2020.

\bibitem{grill2020bootstrap}
Jean-Bastien Grill, Florian Strub, Florent Altch{\'e}, Corentin Tallec,
  Pierre~H Richemond, Elena Buchatskaya, Carl Doersch, Bernardo~Avila Pires,
  Zhaohan~Daniel Guo, Mohammad~Gheshlaghi Azar, et~al.
\newblock Bootstrap your own latent: A new approach to self-supervised
  learning.
\newblock In {\em NeurIPS}, 2020.

\bibitem{guo2020pbl}
Zhaohan~Daniel Guo, Bernardo~Avila Pires, Bilal Piot, Jean-Bastien Grill,
  Florent Altch{\'e}, R{\'e}mi Munos, and Mohammad~Gheshlaghi Azar.
\newblock Bootstrap latent-predictive representations for multitask
  reinforcement learning.
\newblock In {\em International Conference on Machine Learning}, pages
  3875--3886. PMLR, 2020.

\bibitem{MEMDPC}
Tengda Han, Weidi Xie, and Andrew Zisserman.
\newblock Memory-augmented dense predictive coding for video representation
  learning.
\newblock In {\em ECCV}, 2020.

\bibitem{Han20}
Tengda Han, Weidi Xie, and Andrew Zisserman.
\newblock Self-supervised co-training for video representation learning.
\newblock In {\em NeurIPS}, 2020.

\bibitem{he2020momentum}
Kaiming He, Haoqi Fan, Yuxin Wu, Saining Xie, and Ross Girshick.
\newblock Momentum contrast for unsupervised visual representation learning.
\newblock In {\em CVPR}, 2020.

\bibitem{he2015delving}
Kaiming He, Xiangyu Zhang, Shaoqing Ren, and Jian Sun.
\newblock Delving deep into rectifiers: Surpassing human-level performance on
  imagenet classification.
\newblock In {\em ICCV}, 2015.

\bibitem{he16resnet}
Kaiming He, Xiangyu Zhang, Shaoqing Ren, and Jian Sun.
\newblock {Deep Residual Learning for Image Recognition}.
\newblock In {\em {CVPR}}, 2016.

\bibitem{henaff2019data}
Olivier~J H{\'e}naff, Ali Razavi, Carl Doersch, SM Eslami, and Aaron van~den
  Oord.
\newblock Data-efficient image recognition with contrastive predictive coding.
\newblock In {\em ICML}, 2020.

\bibitem{hjelm2018learning}
R~Devon Hjelm, Alex Fedorov, Samuel Lavoie-Marchildon, Karan Grewal, Phil
  Bachman, Adam Trischler, and Yoshua Bengio.
\newblock Learning deep representations by mutual information estimation and
  maximization.
\newblock {\em arXiv preprint arXiv:1808.06670}, 2018.

\bibitem{huang2019unsupervised}
Jiabo Huang, Qi Dong, Shaogang Gong, and Xiatian Zhu.
\newblock Unsupervised deep learning by neighbourhood discovery.
\newblock In {\em ICML}, 2019.

\bibitem{MOCOv2}
Rishabh Jain, Haoqi Fan, Ross~B. Girshick, and Kaiming He.
\newblock Improved baselines with momentum contrastive learning.
\newblock {\em arXiv preprint arXiv:2003.04297}, 2020.

\bibitem{jansen2020coincidence}
Aren Jansen, Daniel~PW Ellis, Shawn Hershey, R~Channing Moore, Manoj Plakal,
  Ashok~C Popat, and Rif~A Saurous.
\newblock Coincidence, categorization, and consolidation: Learning to recognize
  sounds with minimal supervision.
\newblock In {\em ICASSP}, 2020.

\bibitem{jansen2018unsupervised}
Aren Jansen, Manoj Plakal, Ratheet Pandya, Daniel~PW Ellis, Shawn Hershey,
  Jiayang Liu, R~Channing Moore, and Rif~A Saurous.
\newblock Unsupervised learning of semantic audio representations.
\newblock In {\em ICASSP}, 2018.

\bibitem{kay2017kinetics}
Will Kay, Joao Carreira, Karen Simonyan, Brian Zhang, Chloe Hillier, Sudheendra
  Vijayanarasimhan, Fabio Viola, Tim Green, Trevor Back, Paul Natsev, Mustafa
  Suleyman, and Andrew Zisserman.
\newblock The kinetics human action video dataset.
\newblock {\em arXiv preprint arXiv:1705.06950}, 2017.

\bibitem{kingma15adam}
Diederik~P. Kingma and Jimmy Ba.
\newblock Adam: A method for stochastic optimization.
\newblock In {\em ICLR}, 2015.

\bibitem{kong2020panns}
Qiuqiang Kong, Yin Cao, Turab Iqbal, Yuxuan Wang, Wenwu Wang, and Mark~D
  Plumbley.
\newblock Panns: Large-scale pretrained audio neural networks for audio pattern
  recognition.
\newblock In {\em IEEE/ACM Transactions on Audio, Speech, and Language
  Processing}, 2020.

\bibitem{korbar2018cooperative}
Bruno Korbar, Du Tran, and Lorenzo Torresani.
\newblock Cooperative learning of audio and video models from self-supervised
  synchronization.
\newblock In {\em NeurIPS}, 2018.

\bibitem{kuehne2011hmdb}
Hildegard Kuehne, Hueihan Jhuang, Est{\'\i}baliz Garrote, Tomaso Poggio, and
  Thomas Serre.
\newblock {HMDB}: {A} large video database for human motion recognition.
\newblock In {\em ICCV}, 2011.

\bibitem{le2020contrastive}
Phuc~H Le-Khac, Graham Healy, and Alan~F Smeaton.
\newblock Contrastive representation learning: A framework and review.
\newblock {\em IEEE Access}, 2020.

\bibitem{Lee2020Network}
Kimin Lee, Kibok Lee, Jinwoo Shin, and Honglak Lee.
\newblock Network randomization: A simple technique for generalization in deep
  reinforcement learning.
\newblock In {\em ICLR}, 2020.

\bibitem{lin2019tsm}
Ji Lin, Chuang Gan, and Song Han.
\newblock {TSM}: {Temporal} shift module for efficient video understanding.
\newblock In {\em ICCV}, 2019.

\bibitem{loshchilov2017decoupled}
Ilya Loshchilov and Frank Hutter.
\newblock Decoupled weight decay regularization.
\newblock {\em arXiv preprint arXiv:1711.05101}, 2017.

\bibitem{loshchilov2016sgdr}
Ilya Loshchilov and Frank Hutter.
\newblock {SGDR}: {Stochastic} gradient descent with warm restarts.
\newblock In {\em ICLR}, 2017.

\bibitem{malmaud15what}
Jonathan Malmaud, Jonathan Huang, Vivek Rathod, Nick Johnston, Andrew
  Rabinovich, and Kevin Murphy.
\newblock What's cookin'? {Interpreting} cooking videos using text, speech and
  vision.
\newblock {\em NAACL}, 2015.

\bibitem{mathieu2015deep}
Michael Mathieu, Camille Couprie, and Yann LeCun.
\newblock Deep multi-scale video prediction beyond mean square error.
\newblock In {\em ICLR}, 2016.

\bibitem{miech2019end2end}
Antoine Miech, Jean-Baptiste Alayrac, Lucas Smaira, Ivan Laptev, Josef Sivic,
  and Andrew Zisserman.
\newblock {E}nd-to-{E}nd {L}earning of {V}isual {R}epresentations from
  {U}ncurated {I}nstructional {V}ideos.
\newblock In {\em CVPR}, 2020.

\bibitem{miech19howto100m}
Antoine Miech, Dimitri Zhukov, Jean-Baptiste Alayrac, Makarand Tapaswi, Ivan
  Laptev, and Josef Sivic.
\newblock Howto100{M}: Learning a text-video embedding by watching hundred
  million narrated video clips.
\newblock In {\em ICCV}, 2019.

\bibitem{PIRL}
Ishan Misra and Laurens van~der Maaten.
\newblock Self-supervised learning of pretext-invariant representations.
\newblock In {\em CVPR}, 2020.

\bibitem{misra2016shuffle}
Ishan Misra, C~Lawrence Zitnick, and Martial Hebert.
\newblock Shuffle and learn: {Unsupervised} learning using temporal order
  verification.
\newblock In {\em ECCV}, 2016.

\bibitem{morgado20avid}
Pedro Morgado, Nuno Vasconcelos, and Ishan Misra.
\newblock Audio-visual instance discrimination with cross-modal agreement.
\newblock {\em arXiv preprint arXiv:2004.12943}, 2020.

\bibitem{noroozi2016unsupervised}
Mehdi Noroozi and Paolo Favaro.
\newblock Unsupervised learning of visual representations by solving jigsaw
  puzzles.
\newblock In {\em ECCV}, 2016.

\bibitem{oord2018representation}
Aaron van~den Oord, Yazhe Li, and Oriol Vinyals.
\newblock Representation learning with contrastive predictive coding.
\newblock {\em arXiv preprint arXiv:1807.03748}, 2018.

\bibitem{owens2018audio}
Andrew Owens and Alexei~A Efros.
\newblock Audio-visual scene analysis with self-supervised multisensory
  features.
\newblock In {\em ECCV}, 2018.

\bibitem{pathak2016context}
Deepak Pathak, Philipp Krahenbuhl, Jeff Donahue, Trevor Darrell, and Alexei~A.
  Efros.
\newblock Context encoders: Feature learning by inpainting.
\newblock In {\em CVPR}, 2016.

\bibitem{PatrauceanHC16}
Viorica P{\u a}tr{\u a}ucean, Ankur Handa, and Roberto Cipolla.
\newblock Spatio-temporal video autoencoder with differentiable memory.
\newblock In {\em ICLR (Workshop)}, 2016.

\bibitem{mandela2020datatrans}
Mandela Patrick, Yuki~M. Asano, Ruth Fong, João~F. Henriques, Geoffrey Zweig,
  and Andrea Vedaldi.
\newblock Multi-modal self-supervision from generalized data transformations.
\newblock {\em arXiv preprint arXiv:2003.04298}, 2020.

\bibitem{pedregosa2011scikit}
F. Pedregosa, G. Varoquaux, A. Gramfort, V. Michel, B. Thirion, O. Grisel, M.
  Blondel, P. Prettenhofer, R. Weiss, V. Dubourg, J. Vanderplas, A. Passos, D.
  Cournapeau, M. Brucher, M. Perrot, and E. Duchesnay.
\newblock Scikit-learn: Machine learning in {P}ython.
\newblock {\em Journal of Machine Learning Research}, 12:2825--2830, 2011.

\bibitem{piczak2015dataset}
Karol~J Piczak.
\newblock {ESC: Dataset for environmental sound classification}.
\newblock In {\em ACM Multimedia}, 2015.

\bibitem{piergiovanni2020evolving}
AJ Piergiovanni, Anelia Angelova, and Michael~S Ryoo.
\newblock Evolving losses for unsupervised video representation learning.
\newblock In {\em CVPR}, 2020.

\bibitem{qian2020spatiotemporal}
Rui Qian, Tianjian Meng, Boqing Gong, Ming-Hsuan Yang, Huisheng Wang, Serge
  Belongie, and Yin Cui.
\newblock Spatiotemporal contrastive video representation learning.
\newblock {\em arXiv preprint arXiv:2008.03800}, 2020.

\bibitem{richemond2020byol}
Pierre~H Richemond, Jean-Bastien Grill, Florent Altch{\'e}, Corentin Tallec,
  Florian Strub, Andrew Brock, Samuel Smith, Soham De, Razvan Pascanu, Bilal
  Piot, et~al.
\newblock Byol works even without batch statistics.
\newblock In {\em NeurIPS (SSL Workshop)}, 2020.

\bibitem{imagenet}
Olga Russakovsky, Jia Deng, Hao Su, Jonathan Krause, Sanjeev Satheesh, Sean Ma,
  Zhiheng Huang, Andrej Karpathy, Aditya Khosla, Michael Bernstein,
  Alexander~C. Berg, and Li Fei-Fei.
\newblock {ImageNet Large Scale Visual Recognition Challenge}.
\newblock {\em IJCV}, 2015.

\bibitem{Sener_2015_ICCV}
Ozan Sener, Amir~R. Zamir, Silvio Savarese, and Ashutosh Saxena.
\newblock Unsupervised semantic parsing of video collections.
\newblock In {\em ICCV}, 2015.

\bibitem{Senocak_2018_CVPR}
Arda Senocak, Tae-Hyun Oh, Junsik Kim, Ming-Hsuan Yang, and In~So Kweon.
\newblock Learning to localize sound source in visual scenes.
\newblock In {\em Proceedings of the IEEE Conference on Computer Vision and
  Pattern Recognition (CVPR)}, June 2018.

\bibitem{sigurdsson2020visual}
Gunnar~A Sigurdsson, Jean-Baptiste Alayrac, Aida Nematzadeh, Lucas Smaira,
  Mateusz Malinowski, Jo{\~a}o Carreira, Phil Blunsom, and Andrew Zisserman.
\newblock Visual grounding in video for unsupervised word translation.
\newblock In {\em Computer Vision and Pattern Recognition (CVPR)}, 2020.

\bibitem{simonyan2014}
Karen Simonyan and Andrew Zisserman.
\newblock Two-stream convolutional networks for action recognition in videos.
\newblock In {\em ICLR}, 2014.

\bibitem{soomro2012}
Khurram Soomro, Amir~Roshan Zamir, and Mubarak Shah.
\newblock {UCF101}: {A} dataset of 101 human actions classes from videos in the
  wild.
\newblock {\em arXiv preprint arXiv:1212.0402}, 2012.

\bibitem{pmlr-v37-srivastava15}
Nitish Srivastava, Elman Mansimov, and Ruslan Salakhudinov.
\newblock Unsupervised learning of video representations using lstms.
\newblock In {\em ICML}, 2015.

\bibitem{Stroud18}
Jonathan Stroud, David Ross, Chen Sun, Jia Deng, and Rahul Sukthankar.
\newblock D3d: Distilled 3d networks for video action recognition.
\newblock In {\em WACV}, 2020.

\bibitem{sun2019videobert}
Chen Sun, Austin Myers, Carl Vondrick, Kevin Murphy, and Cordelia Schmid.
\newblock {VideoBERT}: {A} joint model for video and language representation
  learning.
\newblock In {\em ICCV}, 2019.

\bibitem{tian2017deepcluster}
Kai Tian, Shuigeng Zhou, and Jihong Guan.
\newblock Deepcluster: A general clustering framework based on deep learning.
\newblock In {\em ECML/PKDD}, 2017.

\bibitem{tian2019contrastive}
Yonglong Tian, Dilip Krishnan, and Phillip Isola.
\newblock Contrastive multiview coding.
\newblock {\em arXiv preprint arXiv:1906.05849}, 2019.

\bibitem{tian2020makes}
Yonglong Tian, Chen Sun, Ben Poole, Dilip Krishnan, Cordelia Schmid, and
  Phillip Isola.
\newblock What makes for good views for contrastive learning.
\newblock In {\em NeurIPS}, 2020.

\bibitem{tran2018closer}
Du Tran, Heng Wang, Lorenzo Torresani, Jamie Ray, Yann LeCun, and Manohar
  Paluri.
\newblock A closer look at spatiotemporal convolutions for action recognition.
\newblock In {\em CVPR}, 2018.

\bibitem{Vondrick16a}
Carl Vondrick, Hamed Pirsiavash, and Antonio Torralba.
\newblock Generating videos with scene dynamics.
\newblock In {\em NIPS}, 2016.

\bibitem{vondrick2018tracking}
Carl Vondrick, Abhinav Shrivastava, Alireza Fathi, Sergio Guadarrama, and Kevin
  Murphy.
\newblock Tracking emerges by colorizing videos.
\newblock In {\em ECCV}, 2018.

\bibitem{walker2016uncertain}
Jacob Walker, Carl Doersch, Abhinav Gupta, and Martial Hebert.
\newblock An uncertain future: Forecasting from static images using variational
  autoencoders.
\newblock In {\em ECCV}, 2016.

\bibitem{Wei_2018_CVPR}
Donglai Wei, Joseph~J. Lim, Andrew Zisserman, and William~T. Freeman.
\newblock Learning and using the arrow of time.
\newblock In {\em CVPR}, 2018.

\bibitem{xie2016unsupervised}
Junyuan Xie, Ross Girshick, and Ali Farhadi.
\newblock Unsupervised deep embedding for clustering analysis.
\newblock In {\em ICML}, 2016.

\bibitem{xie2018rethinking}
Saining Xie, Chen Sun, Jonathan Huang, Zhuowen Tu, and Kevin Murphy.
\newblock Rethinking spatiotemporal feature learning: Speed-accuracy trade-offs
  in video classification.
\newblock In {\em ECCV}, 2018.

\bibitem{xu2020robust}
Zhenlin Xu, Deyi Liu, Junlin Yang, and Marc Niethammer.
\newblock Robust and generalizable visual representation learning via random
  convolutions.
\newblock {\em arXiv preprint arXiv:2007.13003}, 2020.

\bibitem{you2018imagenet}
Yang You, Zhao Zhang, Cho-Jui Hsieh, James Demmel, and Kurt Keutzer.
\newblock Imagenet training in minutes.
\newblock In {\em Proceedings of the 47th International Conference on Parallel
  Processing}, pages 1--10, 2018.

\bibitem{zach2007duality}
Christopher Zach, Thomas Pock, and Horst Bischof.
\newblock A duality based approach for realtime tv-l 1 optical flow.
\newblock In {\em Joint pattern recognition symposium}, 2007.

\bibitem{zhang2016colorful}
Richard Zhang, Phillip Isola, and Alexei~A Efros.
\newblock Colorful image colorization.
\newblock In {\em ECCV}, 2016.

\end{thebibliography}
}

\clearpage
\pagebreak

\appendix

\section*{Appendix}
In this appendix, we provide additional details useful for reproduction of the results. 
In Section~\ref{sec:pretraining_details} we present the details of our training pipeline, including architecture and hyperparameter details (\ref{sec:hyperparameters}), data augmentation and feature extraction (\ref{sec:augmentation}). In Section~\ref{sec:downstream_tasks} we detail the linear and fine-tuning evaluation procedures. Section~\ref{sec:sync} evaluates the importance of syncing video and audio. Section~\ref{sec:image_models} trains \algo with an image narrow backbone and evaluates it on both video and image tasks. Section~\ref{sec:eval_broad} evaluates the broad visual backbone. Finally, Section~\ref{sec:three_broad} discusses the possibility of training \algo with both RGB and flow broad views at the same time. 

\section{Pre-training details}
\label{sec:pretraining_details}

\subsection{Architecture and model hyperparameters}
\label{sec:hyperparameters}

Both predictor and projector are two-layer MLPs with hidden dimensions of $4096$. We use a batch normalisation layer after each hidden layer and after the last layer of the projector. To train our models, we use the LARS~\cite{you2018imagenet} optimizer~\cite{loshchilov2017decoupled} with cosine decay on the learning rate, with $5000$ steps of linear warm up (starting from $0.0$ to the initial learning rate value).  
Models with audio are trained with initial learning rate $1.0$ while models without audio are trained with initial learning rate $4.8$.
We use batch size $512$ and weight decay with value $0.01$, except for R(2+1)d-18 where we use weight decay with value $10^{-7}$.
Following~\cite{grill2020bootstrap}, we multiply the learning rate of the predictor MLPs by $10$.
We train all models for $300$k steps except for the models with audio reported in Table 5, which are trained for $620$k steps. 
We use 16 Cloud TPUs to train all models except for the TSM-50x2 and the R(2+1)D-50 which we train with 32 Cloud TPUs. 
\noindent \subsection{Data augmentation and feature extraction}
\label{sec:augmentation}
\noindent \textbf{RGB}: 
Unless stated othwerwise, we subsample training videos to $12.5$ FPS. For the broad views of the visual only models and the narrow view of the ablation using a $10$s narrow view (first row, Table 1), we subsample training videos to $6.25$ FPS. We also use $6.25$ FPS to subsample the videos when using the R3D architecture.
In terms of spatial augmentations, we use random cropping, random flipping, color jittering, scale jittering and gaussian blurring; sampling their parameters independently for each view. 
Given the original frame, cropping is performed by sampling a bounding box with aspect ratio ranging between $\frac{1}{2}$ and $2.0$ and area between $30\%$ and $100\%$ of the full image. This bounding box is used to crop all frames of the video consistently in time. We horizontally flip all the frames with probability $0.5$.  
With probability of $0.8$ we apply color randomization in brightness, saturation, contrast and hue. This is done by adjusting brightness and hue by an additive offset, each uniformly sampled in respectively $[-32/255, 32/255]$ and $[-0.2, 0.2]$ on a per-sample basis; and similarly, adjusting contrast and saturation by a multiplicative factor, each sampled in $[0.6, 1.4]$.
After this preprocessing, we clip the pixel values in the range $[0, 1.0]$. Furthermore, with probability $0.2$, we convert the RGB sequence to a grayscale sequence. 
Finally, we apply gaussian blur with standard deviation $\sigma$ uniformly sampled in $[0.1, 2.0]$ and with kernel size equal to $\frac{1}{10}$th of the crop side.

\noindent \textbf{Flow}: 
Temporal sampling of the flow is performed similarly to the RGB case for the broad view.
In terms of spatial augmentations, we use random cropping, sampling the crop independently from the narrow view. We also horizontally flip all the frames with probability $0.5$. 
We resize the shortest size of the original frame to 128 and uniformly sample a $112\times 112$ crop. 
We find that scale jittering in flow does not improve performance; as a result, we do not employ this augmentation.

\noindent \textbf{Random Convolutions}:
\label{subsec:randomconv}
Following~\cite{xu2020robust}, we use He initialization~\cite{he2015delving} for the weights, fixed zero bias and dimension-preserving padding. 
We sample the size of the kernel uniformly across odd values ranging from 1 to 11. 
All sampling of kernel size and weights is performed on a per-sample basis.
We refer the reader to the original paper for further details and illustration of the augmentation procedure.

\noindent \textbf{Spectrograms}: The audio is sampled at 48k Hz. We take 80 bins of log-mel spectrograms extracted with Hanning windows of size 320 (6.67 ms) at a stride of 160 (3.33 ms).

\begin{table*}[t]
		\centering
		\caption{\small Hyperparameters for finetuning on HMDB51 and UCF101. \label{tab:hp_ft}}
		\vspace{0.5em}
		\tablestyle{2pt}{1.05}
		\resizebox{0.8\linewidth}{!}{
\begin{tabular}{@{}llrcccccc@{}}
\toprule
\multirow{2}{*}{Method}         & \multirow{2}{*}{Backbone} & \multirow{2}{*}{Dataset} & \multicolumn{3}{c}{HMDB51}       & \multicolumn{3}{c}{UCF101}       \\
                                &                           &                          & Dropout & LR base & Weight decay & Dropout & LR base & Weight decay \\ \midrule
\algo:V$\leftrightarrow$V$\times 3$  & TSM-50      &  K600 & $0.5$          & $0.03$     &  $10^{-7}$ &   $0.1$          & $0.1$     &  $10^{-7}$     \\
\algo:V$\leftrightarrow$F$\times 3$  & TSM-50       &  K600 & $0.5$          & $0.03$     &  $10^{-7}$ &   $0.5$          & $0.1$     &  $0.0$     \\
\algo:V$\leftrightarrow$V$\times 3$  & R3D50      &  K600 & $0.5$          & $0.3$     &  $0.0$ &   $0.1$          & $0.1$     &  $10^{-7}$     \\
\algo:V$\leftrightarrow$F$\times 3$  & R3D50       &  K600 & $0.8$          & $0.03$     &  $0.0$ &   $0.1$          & $0.3$     &  $10^{-7}$     \\
\algo:V$\leftrightarrow$V$\times 3$  & R3D50      &  K400 & $0.1$          & $0.3$     &  $0.0$ &   $0.1$          & $0.3$     &  $10^{-7}$     \\
\algo:V$\leftrightarrow$F$\times 3$  & R3D50       &  K400 & $0.5$          & $0.03$     &  $10^{-7}$ &   $0.1$          & $0.3$     &  $10^{-7}$     \\
\algo:V$\leftrightarrow$A  & R(2+1)D-18  &  AS   & $0.8$          & $0.3$     &  $10^{-7}$ &   $0.8$          & $0.1$     &  $10^{-7}$      \\
\algo:V$\leftrightarrow$A  & TSM-50      &  AS   & $0.8$          & $0.03$     &  $10^{-7}$ &   $0.5$          & $0.1$     &  $0.0$     \\
\algo:V$\leftrightarrow$FA & TSM-50     &  AS   & $0.8$          & $0.03$     &  $10^{-7}$ &   $0.5$          & $0.1$     &  $0.0$     \\
\algo:V$\leftrightarrow$FA & R(2+1)D-50 &  AS   & $0.5$          & $0.015$     &  $10^{-7}$ &   $0.5$          & $0.15$     &  $0.0$     \\
\algo:V$\leftrightarrow$FA & TSM-50x2   &  AS   & $0.8$          & $0.05$     &  $10^{-7}$ &   $0.1$          & $0.15$     &  $10^{-7}$    \\ \bottomrule
\end{tabular}
}
\end{table*}
\section{Downstream task evaluation}
\label{sec:downstream_tasks}
\paragraph{Linear evaluation on HMDB51, UCF101 and ESC-50.}
For the linear evaluation on HMDB51, UCF101, and ESC-50 we use the SVM implementation of SciKit-Learn~\cite{pedregosa2011scikit}. 
For all three datasets, we use the same augmentations as during the pre-training stage except for gaussian blurring, and process $10$ epochs worth of augmented samples.
For each sample, we extract features using the pre-trained backbone.
We find it helpful to rescale the features using a batch norm layer with fixed scaling and offset parameters (respectively of 1 and 0), collecting training statistics over the extracted features.
We sweep the value for the regularization parameter of the SVM in the following set of values: $\{10^{-5}, 3\cdot 10^{-5}, 10^{-4}, 3\cdot 10^{-4}, 10^{-3}, 3\cdot 10^{-3}, 10^{-2}\}$. 
For all models and downstream tasks, we use the first split to pick the optimal value and report the average of all the splits in that regime. 
At test time, we do not apply any augmentation. 
We subsample test videos to the original FPS used when training the narrow backbone. For HMDB51 and UCF101, given a test video, we resize the minimum side to $256$ and then average the predictions over 30 clips of size $224 \times 224$ (10 temporal clips regularly spaced within the video each providing 3 spatial crops). For HMDB51 and UCF, we use clips of $32$ frames.
For ESC-50 we use a single window of $5$s at test time. 
Finally, one special case is the ablation with a $10$s narrow view (first row, Table 1), which is trained with $64$ frames at $6.25$ FPS and $112 \times 112$ crops. For fairness, we evaluate it with clips of size $112 \times 112$ (minimum side $128$) of $64$ frames subsampled at $6.25$ FPS (same frame rate than in training).

\paragraph{Finetuning evaluation on HMDB51 and UCF101.}
For fine-tuning, we use the same optimiser as when pre-training, the LARS~\cite{you2018imagenet} optimizer. As during pre-training, we do not apply LARS to biases and batch norm parameters.
We use a batch size of $256$ for all methods except for TSM-50x2 and R(2+1)D-50 for which we use a smaller batch size of $128$.
The batch is distributed over 32 workers.
Although we use cross replica batch norm during pre-training (\ie the statistics are accumulated over the 32 workers), during finetuning, we find it better to only compute statistics of batch norm within each worker.
We hypothesize that this has a regularization effect on these small datasets.
We use a linear warm up for the learning rate for 50 epochs (starting from $0.0$ to the initial learning rate value).
Learning rate is then decreased using a cosine decay for 550 epochs.
Weight decay is employed on the weights of the network (except bias and batch norm parameters).
We also apply dropout before the last linear layer mapping the representation to the logits of the classes. Furthermore, we use a batch normalisation layer before the linear layer which we find useful to improve performance.
We cross validate the value of the initial learning rate (taking values in $\{0.03, 0.1, 0.3\}$), the weight decay (taking values in $\{0., 10^{-7} \}$) and dropout rate (taking values in $\{0.1, 0.5, 0.8\}$). For models using batch size $128$, we divide the values of the learning rate by 2 (taking values in $\{0.015, 0.05, 0.15\}$).
Similarly to the linear setting, we select hyperparameters on split 1 of each downstream task and report averaged performance values across splits.
The values of hyperparameters found for all networks are given in Table~\ref{tab:hp_ft}.
For training, we apply the following augmentation procedure in this order: temporal sampling, scale jittering, resizing the minimum side to $256$, extracting a random crop of $224\times224$ and  random horizontal flipping.
For temporal sampling, we randomly sample in time a subclip of 32 frames from the original video clip.
For scale jittering, we independently scale width and height by a value uniformly sampled from $[0.8, 1.2]$.
At test time, we resize the minimum side to $256$ and then average the predictions over 30 clips of size $224 \times 224$ (10 temporal clips regularly spaced within the video each providing 3 spatial crops).
We use the same FPS as during pre-training for each model.

\paragraph{Linear evaluation on Kinetics-600.}
Since Kinetics-600 is too large to fit in memory, we cannot use Scikit-Learn directly.
Instead we train the linear layer for $100$ epochs using the LARS~\cite{you2018imagenet} optimizer with a batch size of 512. As during pre-training, we do not apply LARS to biases and batch norm parameters.
We found it beneficial to apply batch norm before the linear layer.
We use a linear warm up for the learning rate for $6.5$ epochs (starting from $0.0$ to the initial learning rate value).
We use an initial learning rate of $0.5$ for all models. We do not use any weight decay. 
For training, we apply the same augmentations as in pre-training except for the gaussian blur.
For temporal sampling, we randomly sample in time a subclip of 32 frames from the original video clip.
At test time, we resize the minimum side to $256$ and then average the prediction over 30 clips of size $256 \times 256$ (10 temporal clips linearly spaced within the video each with 3 spatial crops).
We do not apply scale jittering or horizontal flipping during test time.
We use the same FPS as during pre-training.
We report the top 1 accuracy on the validation set of Kinetics-600.

\paragraph{Shallow classifier evaluation on AudioSet.} Following the protocols in \cite{jansen2018unsupervised, jansen2020coincidence, alayrac2020self}, we evaluate our audio representations by training a shallow MLP on AudioSet. The MLP has 1 hidden layer with 512 units, and is trained with the Adam optimizer using a batch size of 512 for 20 epochs. We use batch normalization layers on the frozen audio features and after the hidden layer. A ReLU activation function is applied after the second batch normalization. We use a linear warm up of 5000 steps starting from $0.0$ to the initial learning rate of $2\times10^{-4}$, which then decays following a cosine function. At test time, we use 10 overlapping crops of length 5s regularly spaced throughout the audio clip.

	\begin{table}[t]
		\centering
		\caption{\small {\bf Sync study.} Effect of syncing the narrow and broad views.
        }
        \vspace{0.5em}
		\resizebox{\linewidth}{!}{
		\begin{tabular}{ccc|ccc} \toprule
			Dataset & Sync  & $M_b$ & HMDB51 &UCF101& K600 \\ 
			\midrule
 			AS & Async & Audio & 67.4   & 92.3  & 69.1  \\
 			AS & Center & Audio &  67.3 &  92.2 & \bf 70.6  \\
 			AS & Start & Audio & \bf 68.2 & \bf 92.7 & 69.7  \\

			\bottomrule
		\end{tabular}
	}
		\label{tab:sync_apendix}
	\end{table}

\section{Syncing audio and video}
\label{sec:sync}
Table~\ref{tab:sync_apendix} shows the performance of a model trained with broad audio view when the narrow view and the broad view start at the same temporal instant (\emph{start sync}), the narrow view is centered on the broad view (\emph{center sync}) or are independently randomly sampled in time (\emph{async}). As discussed in Section~\ref{subsect:framework} in the paper, this experiment supports already established evidence~\cite{korbar2018cooperative} that syncing audio and video is beneficial for the resulting model in self-supervised learning. Overall, we find syncing at the beginning the best option.

\section{Using \algo to learn image models }
\label{sec:image_models}
In this section we explore learning image models using \algo. For that, we train two models. Both models use a 1 frame narrow view (which can be used as an image model). One model has a $10$s broad visual view (RGB+RC) and the other one has a $5s$ audio broad view.
Although the task is particularly challenging (regress a representation of the broad video from a single frame) the resulting model is able to learn strong image and video features.
When evaluated in Imagenet linear the visual-trained model scores $56.4$ while the model self-supervised with audio scores $59.2$. Furthermore, when evaluated in Kinetics-600 linear the visual-trained model has an accuracy of $60.5$ while the model using an audio broad view perfoms at $60.8$. Those results show that \algo can learn descriptive image  and video representation. We did not see any performance improvement when using \algo as pre-training for the detection task. We attribute this result to the semantic nature of the loss, which does not need to keep spatial information to work.

\section{Evaluating the broad backbone }
\label{sec:eval_broad}
In Table~\ref{tab:sota} we report the performance of the broad audio backbone. However, we are also interested on analysing the performance of the broad backbone when using visual data. Of those, we can only evaluate those using RGB to obtain comparable performance. We evaluated the model trained with RGB broad view (without RC) and obtained a performance of 63.4 in HMDB51 and 90.3 in UCF101, while the narrow backbone performs at 63.2/90.5 respectively. While using the broad network on RGB does indeed perform on par, our best results are obtained when the broad view is from a different (flow or audio) or altered modality (e.g. RC). In that case the comparison is not possible.

	\begin{table}[t]
		\centering
		\caption{\small {\bf Combinations of broad views.}  
		We compare the different possible combinations of visual broad views.
		We use $\tau_n = 1.3s$ (narrow extent) and $\tau_b=10s$ (broad extent).
		RC stands for random convolutions. 
        }
        \vspace{0.5em}
		\begin{tabular}{cc|ccc} \toprule
			RGB+RC & Flow & HMDB51 &UCF101 & K600 \\ 
			\midrule
			\midrule
			  1 & \xmark &65.2 & 91.7 & 69.1  \\
			  2 & \xmark &65.4 & 91.8 & 70.4  \\
			  3 & \xmark & 65.4 & 92.6& \textbf{70.8}  \\
			 \xmark & 1 & 66.4 & 92.1 & 67.0   \\
			  \xmark  & 3 & 66.8& 92.9 & 68.1   \\
			  1  & 1 & \textbf{68.5} & \textbf{93.2} & 69.6  \\

			\bottomrule
		\end{tabular}

		\label{tab:three_broad}
	\end{table}

\section{Training \algo with RGB and flow views }
\label{sec:three_broad}
In the main paper, we focus our analysis on different configurations of \algo using one modality for visual broad view (either RGB or Flow), one modality for the audio broad view or a combination of both. However, it is possible to use multiple visual broad views, each of those being of a different modalities. As opposed to using multiple views of the same modality, this requires an additional backbone model and thus additional computational resources. In this section we evaluate the performance of \algo trained using both RGB and flow broad views. We report the resulting performance in Table~\ref{tab:three_broad}, comparing it with other possible combinations of broad views. Using multiple visual modalities in the broad view results on improvements on HMDB51 and UCF101, but decreases the performance on K600 compared to using 2 RGB views. As shown by the version of \algo using only flow, the K600 benchmark does not benefit from the information contained on that modality as much as other benchmarks. We hypothesise that the increase in performance on HMDB51 and UCF101 is due to the complementary information provided by RGB and flow together with the larger overall capacity of the broad model (two broad backbones). We acknowledge that this version of \algo is not directly comparable to using a single backbone for the broad view, but we believe those results are informative to gain intuition about the model.
\end{document}